\documentclass{article}
\usepackage{graphicx} 
\usepackage{amsfonts, mathrsfs, amsmath, amsthm}
\usepackage{enumitem}
\usepackage{appendix}
\usepackage[section]{algorithm}
\usepackage{algpseudocode}
\usepackage{mathtools}

\usepackage{comment}
\usepackage{booktabs} 
\usepackage{diagbox}
\usepackage{tabularx}
\usepackage{minitoc}
\usepackage{todonotes}

\usepackage[parfill]{parskip} 
\usepackage{subcaption}
\usepackage{svg}
\usepackage{xcolor}
\newtheorem{assumption}{Assumption}[section]
\theoremstyle{definition}

\newtheorem{proposition}{Proposition}[section]

\newcommand{\E}{\mathbb{E}}

\newcommand{\la}{\langle}
\newcommand{\ra}{\rangle}

\newcommand*\de{\mathop{}\!\mathrm{d}}
\newcommand*\diver{\mathop{}\!\mathrm{div}}
\newcommand{\dist}{\mathrm{dist}}

\usepackage[normalem]{ulem}

\numberwithin{equation}{section}
\title{Score-based constrained generative modeling via Langevin diffusions with boundary conditions }
\author{%
Adam Nordenhög\thanks{%
Institute for Analysis and Scientific Computing,
TU Wien. \texttt{} \texttt{}}
\and
Akash Sharma\thanks{%
Department of Mathematical Sciences,
Chalmers University of Technology and the University of Gothenburg; \texttt{akashs@chalmers.se}.} 
}
\date{}

\begin{document}

\maketitle
\begin{abstract}
Score-based generative models based on stochastic differential equations (SDEs) achieve impressive performance in sampling from unknown distributions, but often fail to satisfy underlying constraints.
We propose a constrained generative model using kinetic (underdamped) Langevin dynamics with specular reflection of velocity on the boundary defining constraints. This results in piecewise continuously differentiable noising and denoising process where the latter is characterized by a  time-reversed dynamics restricted to a domain with boundary due to specular boundary condition. In addition, we also contribute to existing reflected SDEs based constrained generative models, where the stochastic dynamics is restricted through an abstract local time term. By presenting efficient numerical samplers which converge with optimal rate  in terms of discretizations step, we provide a comprehensive comparison of models based on confined (specularly reflected kinetic) Langevin diffusion with models based on reflected diffusion  with local time.

   \medskip
       
   
   \medskip

\end{abstract}

\section{Introduction}
Generative modeling via diffusion models (see \cite{sohl2015deep, song2019generative, ho2020denoising, song2020sliced, song2021denoising}) is a  powerful approach for synthesizing high-quality data \cite{dhariwal2021diffusion}. This approach relies on three key components: a forward process to noise the data, a loss function to train the neural network, and a reverse dynamics which uses the  learned neural network to iteratively turn the noise into structured outputs.
A powerful framework is provided by stochastic differential equations (SDEs) (see \cite{song2021scorebased}) to identify the time-reversal (see \cite{haussmann1986time}) of the forward (noising) process and then a numerical scheme can be employed to discretize the reverse dynamics to generate new samples.

\begin{figure}[htbp]\label{fig:first_page}
\centering
\includegraphics[width=0.25\textwidth, height = 0.1\textheight]{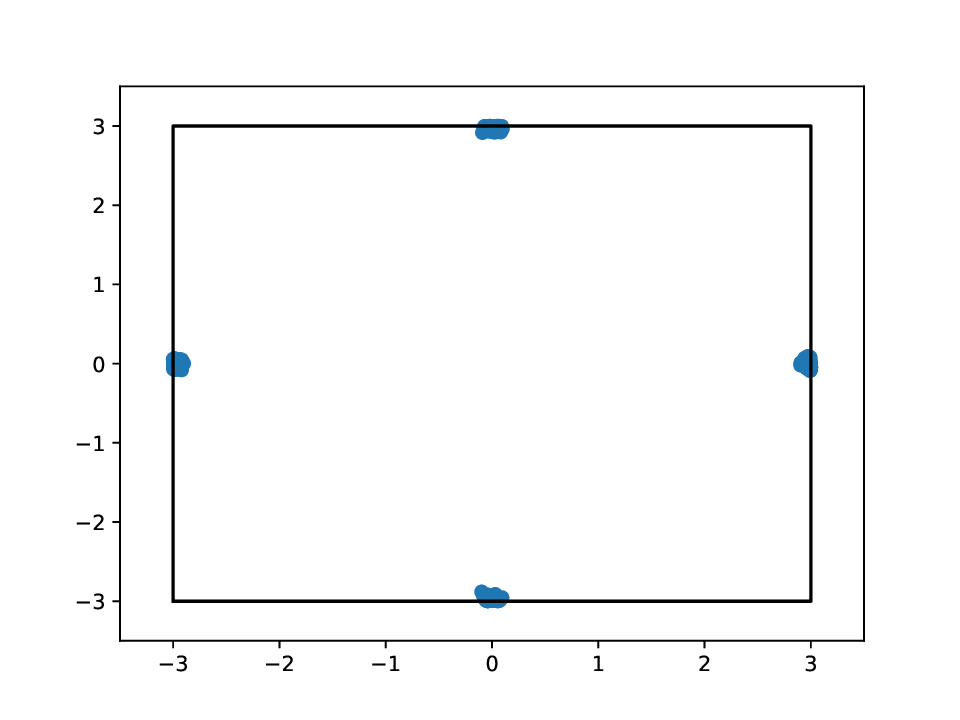}
\hfill
\includegraphics[width=0.25\textwidth, height = 0.1\textheight]{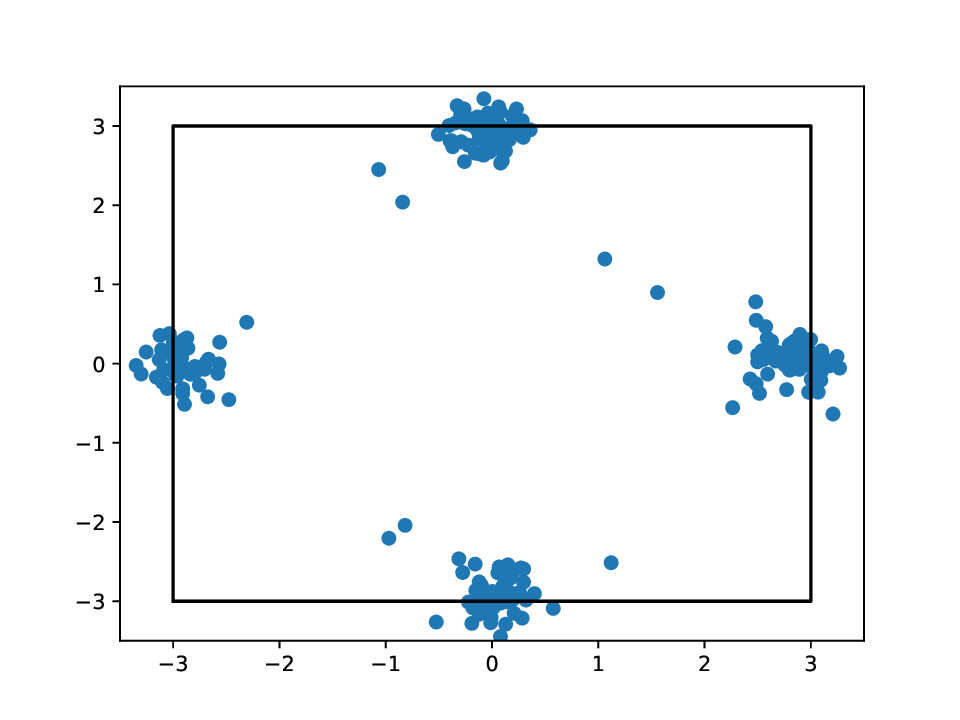}
\hfill
\includegraphics[width=0.25\textwidth, height = 0.1\textheight]{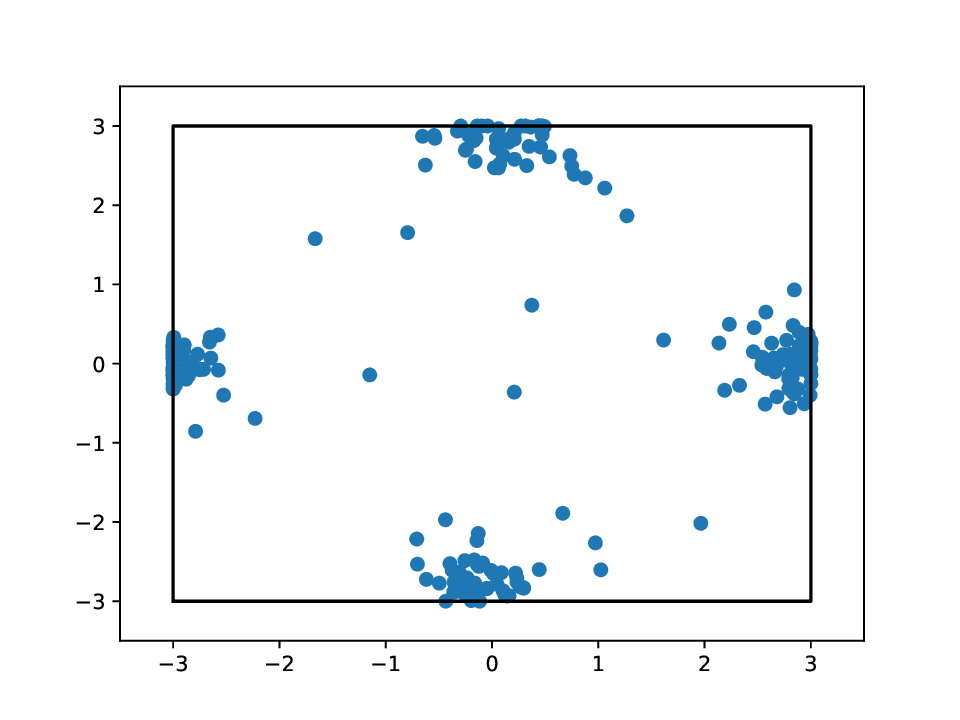}
\caption{A simple comparison : Original data; Data generated by unconstrained diffusion model; Data generated by projected diffusion model. } \label{fig:first_page} 
\end{figure}

 If there are stringent data constraints, although diffusion models will try to imitate the underlying distribution, the final generated output may not respect the constraints due to the injected noise at each step of the sampling process (see Figure~\ref{fig:first_page}). To remedy above mentioned issues, one proposed approach is based on reflected SDEs developed in \cite{lou2023reflected} to incorporate data constraints in  diffusion model. From a theoretical viewpoint, the same reflected model is also considered in \cite{fishman2023diffusion} in manifold setting with some practical differences. In addition, \cite{fishman2023diffusion} also considers a barrier method turning the constraint set into a Reimannian manifold. Another approach that has been introduced in \cite{liu2023mirror}, particularly to deal with convex constraints, is based on mirror map which transforms a constrained space into unconstrained dual space. Mirror diffusion models require solving SDEs in the dual space which is a Hessian manifold, therefore, the sampling process requires the use of an exponential map (or its approximation) based scheme \cite{bharath2025sampling} as compared to Euler-Maruyama scheme. The authors of
 \cite{fishman2023metropolis} consider a Metropoloized random walk which rejects a proposed next step in the sampling process if it violates the constraints.
Projected diffusion model, without using time-reversal, is proposed in \cite{christopher2024constrained}  
replacing Steins score with learned score neural network, to generate samples. This model can also be considered as discretization of reflected Langevin dynamics. It is worth mentioning other generative models proposed in recent works which are either based on reflected (Skorokhod) dynamics, but related to mechanisms involving flow matching or Schrödinger bridge, see for example   \cite{deng2024reflected_Scro_bridge,xie2024reflected_flow, naderiparizi2025constrained_bridge} or posing the constrained generative models as infinite dimensional constrained optimization problem \cite{khalafi2024constrained}.

\paragraph{Notation.}
Constraint set is represented by $G \subset \mathbb{R}^{d}$ and its boundary we denote with $\partial G $. We denote with $n(x)$, the outward unit normal at $x \in \partial G$. Also, $\bar{G}$ represents the closure of $G$ and $\bar{G}^{c}$ denotes the complement of $\bar{G}$. We denote $\la a \cdot b \ra$ as dot product between $a, b \in \mathbb{R}^{d}$ and $|a |$ as Euclidean norm of $a \in \mathbb{R}^{d}$. For a vector valued function $f : \mathbb{R}^d \rightarrow \mathbb{R}^d$, we use the notation $\diver(f)$ as well as $\nabla_x \cdot f$ (according to convenience) for divergence of $f$.  Also, $\mathcal{N}(0,1)$ and $\mathcal{U}([0,T])$ denote standard normal distribution and uniform distribution on $[0,T]$, respectively. We denote with $W_t$ and $B_t$ as Brownian motions in forward and reverse dynamics, respectively.

\paragraph{Our contributions.} 

We present a new diffusion model based on \textbf{confined Langevin dynamics}, whose positional movement is restricted to confined domain via c\'{a}dl\'{a}g (right continuous with left limits) process. This results in strict adherence of constraints without any mismatch in training and generating process.  The dynamics includes a velocity variable, inspired from collisional mechanical systems. In the unconfined setting this has been considered in \cite{dockhorn2022scorebased} showing improved performance thanks to smoother noising and denoising process.   We characterize the time reversal of confined Langevin dynamics and derive loss function to train a score network. 

In addition, we also contribute to the existing work on reflected diffusion models \cite{lou2023reflected, fishman2023diffusion, christopher2024constrained}.
We (computationally) investigate various methods, namely projection, symmetrized reflection, specular reflection, penalty, and barrier (different from \cite{fishman2023diffusion}, see Section~\ref{sec_3.2_nrps_rsde}) methods for handling constraints which are approximations (in a weak or $L^2$ sense) of the local time (term responsible to restrict continuous dynamics to a bounded domain). 
  
  Moreover, we deal with an issue related to loss function identified in \cite{lou2023reflected}. A claim was put forth in \cite[Section~4.1]{lou2023reflected} that implicit score matching is not tractable due to term
$
    \int_{\partial G} \rho_t(x) \la s_\theta(t,x) \cdot n(x) \ra \de x
 $ 
where $\rho_t$ denote the time-marginal distribution.  To address this issue, the authors in \cite{fishman2023diffusion} impose a condition that $s_{\theta}(t,x) = 0$ for $x \in \partial G$. 
 We argue otherwise that the term is tractable and can be approximated with a weighted estimator (see Propositions~\ref{ref_lossProp_2}-\ref{ref_lossProp_3} for more details) without any extra condition on score network.

 Finally, we compare the performance of various numerical samplers corresponding to confined Langevin diffusion model (see Subsection~\ref{subsec_2.1_gencld}) and reflected diffusion model (see Subsection~\ref{subsec_2.2_genrsde}) providing more insights in the implementation and theory of constrained diffusion models.  One drawback of this paper is that we do not have large scale experiments. 

\section{Langevin diffusions with boundary conditions and their time-reversal}
In Subsection~\ref{gen_cld_subsec_review}, we first review the mechanism behind the generative modeling using SDEs. 
In Subsection~\ref{subsec_2.1_gencld}, we introduce the confined Langevin dynamics, Fokker-Planck PDE with specular boundary condition describing the evolution of its time-marginal law,  and time-reversed (in distributional sense) dynamics. In Subsection~\ref{subsec_2.2_genrsde}, following the same order, we present the reflected (overdamped) Langevin dynamics, Fokker-Plank PDE and the time-reversed dynamics.  
We mention in detail the difference between two constrained Langevin diffusions in Table~\ref{comparison_table_1}.

\subsection{Brief overview of SDEs based generative modeling in $\mathbb{R}^d$}\label{gen_cld_subsec_review}

Let $\mu_{\text{data}}$ denotes the data distribution. Assume we have a large number of samples from this distribution. The aim is to generate new samples from the probability measure representing data distribution. 

The fundamental component of diffusion based generative  modeling is the data generation process as the reverse of a noising process, where noising process is responsible of progressively corrupting original data  with noise until it becomes indistinguishable from random noise. In continuous time, this forward noising process can be mathematically described as a stochastic differential equation (SDE) of the form 
\begin{align}\label{cld_neweqn_2.1}
    \de X_t = b(t, X_t) \de t + \sigma(t) \de W_t,\quad X_0 \sim \mu_{\text{data}}, 
\end{align}
 where $b : [0,T] \times \mathbb{R}^{d} \rightarrow \mathbb{R}^d$  is the drift term,  $\sigma : [0, T] \rightarrow \mathbb{R}_{+}$ is the diffusion coefficient and $W_t$ represents Brownian motion. The forward SDE systematically destroys the data structure by adding noise over continuous time $t \in [0,T]$. Let $\rho_t$ denotes the density of the time-marginal law of $X_t$. 
 
The key insight of diffusion models lies in time-reversal theory \cite{anderson1982reverse, elliott1985reverse, haussmann1986time}, which shows that the reverse SDE also exists which is also driven by a Brownian motion. This  can be used to generate data from noise. Here, the time reversal is in terms of law of the dynamics, i.e. $Y_t$ is time reversal of $X_t$ if $Y_t \sim X_{T-t}$ for a fixed $T>0$. According to \cite{anderson1982reverse, elliott1985reverse, haussmann1986time}, the reverse-time SDE takes the form 
\begin{align}
  \de Y_t = [-b(T-t,Y_t) + \sigma^2(T-t) \nabla_x \log \rho_{T-t}(Y_t)]\de t + \sigma(t)\de B_t,  \quad Y_{0} \sim X_{T-t},
\end{align}
 where the crucial intractable term $ \nabla_x \log \rho_{T-t}(Y_t) $ is the score function (gradient of the log probability density, $\rho_t$).

 However, the score function is generally unknown and must be learned from data.   Training diffusion models involves learning a neural network $s_\theta (t, x)$, where $\theta$ represents parameters (weights and biases) of the neural network,  to approximate the unknown score function $\nabla_x \log \rho_t(x)$ at each time step. This is achieved through score matching. The training objective is typically the denoising score matching loss: 
 \begin{align}\label{cld_neweqn_2.3}
      \text{Loss} = \E_{t \sim \mathcal{U}([0,T]),X_0\sim \mu_{\text{data}}} [|s_\theta (t,X_t) - \nabla_x \log \rho_t(X_t)|^2]
 \end{align}
The above is a brief recap of underlying theory of generative diffusion models \cite{song2021denoising, song2021scorebased} via SDEs. 

In \cite{dockhorn2022scorebased}, a generative model based on kinetic Langevin diffusion is proposed. In the score-based generative models \eqref{cld_neweqn_2.1}-\eqref{cld_neweqn_2.3}, the neural network must learn the score function $\nabla_{x} \log p_t$ of the marginal distribution of noisy data, which becomes increasingly complex as the data distribution is progressively corrupted. On the contrary, kinetic Langevin diffusion addresses this limitation by operating in an augmented space $(X_t, V_t)$ where data variables $X_t$ are coupled with auxiliary velocity variables $V_t$.  Consequently, the model only needs to learn the score function of the conditional distribution of the velocity given the data. 

In our notation, the forward dynamics is   
\begin{align}
\de X_t &= V_t \de t, \quad \quad \quad \quad  X_{0} \sim \mu_{\text{data}},\label{kld_1} \\
\de V_t &=  b(X_t) \de t  -  \gamma_t V_t \de t + \sqrt{2\gamma_t} \de W_t,\quad \quad V_{0} \sim \mathcal{N}(0, I), \label{kld_2}
 \end{align}
and the time-reverse dynamics, i.e. $(Q_t, P_t) \sim (X_{T-t}, V_{T-t})$ is 
\begin{align}
    \de Q_t &=  -P_t \de t, \label{rev_kld_1} \\
     \de  P_t & = -  b( Q_t) \de t +  \gamma_{T-t} P_t \de t + 2 \gamma_{T- t}\nabla_p \ln \rho(T-t, Q_t, P_t)\de t \nonumber \\
       & \quad  + \sqrt{2\gamma_{T- t}}\de B_t,  \label{rev_kld_2}
\end{align}
In next section, we introduce a constrained generative model based on \eqref{kld_1}-\eqref{kld_2}. In this model, position variable is constrained to evolve on a bounded subset of $\mathbb{R}^d$.

\subsection{Confined kinetic Langevin diffusion}\label{subsec_2.1_gencld}
Let $\gamma_s : [0,T] \rightarrow \mathbb{R}_{+}$.  Let $(X_t, V_t)$ denote the position and velocity of a Markovian unit mass particle driven by the following SDEs: 

\begin{align}
    X_t &= X_0 + \int_{0}^{t} V_s \de s, \quad\quad\quad   \label{cld_pos_for} \\
      V_t & = V_0 + \int_{0}^t b(s, X_s) \de s -  \int_{0}^t \gamma_s V_s \de s + \int_{0}^t\sqrt{2\gamma_s} \de W_s  -2 \sum_{0< s \leq t} \langle n(X_s) \cdot  V_s\rangle n(X_s)I_{\partial G}(X_s), \label{cld_vel_for}
\end{align}
with $(X_0, V_0) \in G\times \mathbb{R}^{d}$. The particle is restricted to evolve on $G$ due to the specular reflection of the velocity when $X_t$ hits the boundary $\partial G$. The existence and uniqueness of the solution of SDEs has been discussed in \cite{bossy_jabir_2015, spiliopoulos_07, bossy2011conditional}.  One can think of SDEs \eqref{cld_pos_for}-\eqref{cld_vel_for} as composed of three components : (i) deterministic Hamiltonian type dynamics; (ii) Orstein-Uhlenbeck impulse in the velocity component; and (iii) specular reflection of velocity when dynamics hit the boundary $\partial G$.

\begin{figure}[H]
        \centering
        \includegraphics[ width=0.5\linewidth] 
        {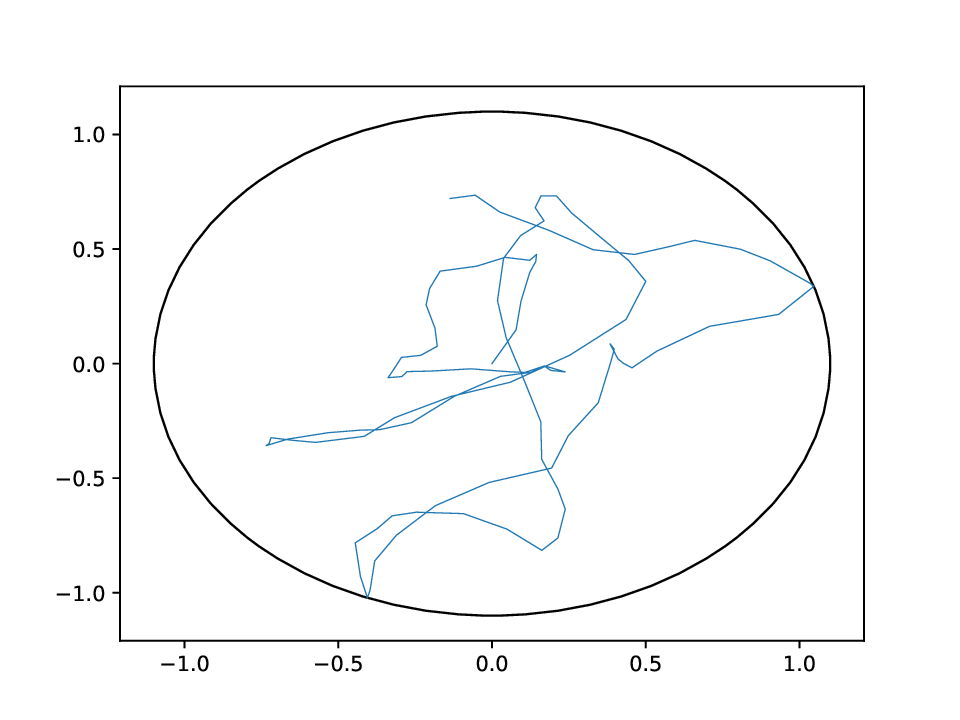}
        \caption{Trajectory of $X_t$ of confined Langevin dynamics \eqref{cld_pos_for}-\eqref{cld_vel_for}.}
\end{figure}
The time-evolution of probability distribution is described by the following partial differential equation (PDE) with specular boundary condition (see \cite{bossy_jabir_2015, bossy2011conditional, ros2025optimal_spec_pde}) : 
\begin{align}
    \frac{\partial \rho_t}{\partial t} &= -\la v \cdot \nabla_x \rho_t\ra  - \la \nabla_v  \rho_t \cdot b(t, x)\ra + \gamma_t  \nabla_v \cdot (v \rho_t) + \gamma_t\Delta_v \rho_t,\;\; (x,v) \in \bar{G}\times \mathbb{R}^d, \label{spec_pde_1} \\
    \rho_t(x, v) &=  \rho_t(x, v - 2\la n(x)\cdot v\ra n(x)),\;\;\; (x,v) \in \{ x\in \partial G\; ; \; \la n(x)\cdot v\ra >0\}.\label{spec_pde_2}
\end{align}

Let us take $\gamma$ to be a constant and fix $b(x) := -\nabla U(x) $ for some scalar valued potential function $U : \bar{G} \rightarrow \mathbb{R}$. Let $\mathcal{G}(\de x)$ denote the Gibbs measure on $G$ i.e. $\mathcal{G}(\de x) \propto e^{-U(x)}\de x$, where $\de x$ denotes Hausdorff measure on $G$. Under appropriate conditions on $U$, we can show the following ergodic result:  
\begin{align}
      \lim_{T\rightarrow \infty}\frac{1}{T}\int_{0}^{T}\varphi(X_s)\de s = \int_{G} \varphi(x) \mathcal{G}(\de x), \; \text{a.s.},
\end{align}
where $\varphi$ belongs to sufficiently large class of functions.  If we take $b(x) \equiv 0$ then the time-marginal law of $X_t$ converges to uniform measure on $G$ if $G$ is bounded.

\paragraph{Time-reversal of confined Langevin dynamics.} Thanks to the Fokker-Planck equation, we can characterize the reverse dynamics as follows: 
\vspace{-5pt}
\begin{align} 
    Q_t &= Q_0 - \int_{0}^{t} P_s \de s, \label{rev_cld_1} \\
      P_t & = P_0 - \int_{0}^t b(T- s, Q_s) \de s +  \int_{0}^t  \gamma_{T-s} P_s \de s + 2\int_{0}^t \gamma_{T- s}\nabla_p \ln \rho(T-s, Q_s, P_s)\de s \nonumber \\
       & \quad  + \int_{0}^t\sqrt{2\gamma_{T- s}}\de B_s  - 2 \sum_{0< s \leq t} \langle n(Q_s) \cdot  P_s\rangle n(Q_s)I_{\partial G}(Q_s),  \label{rev_cld_2}
\end{align}
such that $(Q_t, P_t) \sim (X_{T- t}, V_{T- t})$ for some fixed $T > t \ge 0$. As is the case in other score based models, $\nabla_{v} \ln \rho(t, x, v)$ is replaced by a score network $s_{\theta}(t, x, v)$ \cite{song2019generative} learnt via score matching loss function.

\subsection{Reflected (overdamped) Langevin diffusion}\label{subsec_2.2_genrsde}

Consider the following overdamped Langevin diffusion with reflection \cite{tanaka1979stochastic, lions1984stochastic, saisho1987stochastic}: 
\begin{align}
    X_t = X_0 +\int_{0}^t b(X_s) \de s + \sqrt{2}W_t - \int_{0}^{t} n(X_s) \de L_s, \label{over_damp_Lang_sde}
\end{align}
where $L_s$ is a scalar non-decreasing finite variation process which increases only if $X_s \in \partial G$. The time-marginal probability distribution satisfies the following PDE with Neumann boundary condition (see \cite{miranda2013partial, freidlin_book, bencherif2009probabilistic}):
\begin{align}
\frac{\partial \rho_t}{\partial t} &= - \nabla_x \cdot (\rho_t b(x)) + \Delta_x \rho_t,\quad x \in \bar{G}, \label{neum_pde_1}
\\  
\la b(x) \cdot n(x)\ra \rho_{t} &+ \la\nabla \rho_t \cdot n(x)\ra = 0,\quad x \in \partial G.    \label{neum_pde_2}
\end{align}
Note the distinction between uniformly elliptic Neumann boundary value problem (\ref{neum_pde_1})-(\ref{neum_pde_2}) and hypoelliptic PDE (\ref{spec_pde_1}) with specular boundary condition (\ref{spec_pde_2}) which represents a jump in velocity variable.  If we take $b \equiv 0$ then the law of $X_t$  (from (\ref{over_damp_Lang_sde})) converges  to uniform distribution on bounded $G$ \cite{cattiaux1992stochastic}. 
\paragraph{Time-reversal of reflected diffusion.}
The time-reversal of the reflected SDE, i.e. $Y_t \sim X_{T-t}$, is again a reflected diffusion (see \cite{cattiaux1988time_reversal_boundary, haussmann1986time}) and is given by
\begin{align}
    Y_t = Y_0 + \int_{0}^t (-b(Y_s) + \nabla_y \ln \rho(T-s, Y_s)) \de s + \sqrt{2}B_t + \int_{0}^{t}n(Y_s)\de L^{Y}(s), \label{time_reverse_dyn}
\end{align}
with $Y_{0} \in \bar{G}$. Replacing $\nabla_{x}\ln \rho(t, x)$ in (\ref{time_reverse_dyn}) with learnt score network $s_{\theta}(t, x)$, we get the generating process. 
\begin{figure}[htbp]
        \centering
       \includegraphics[width=0.5\linewidth]{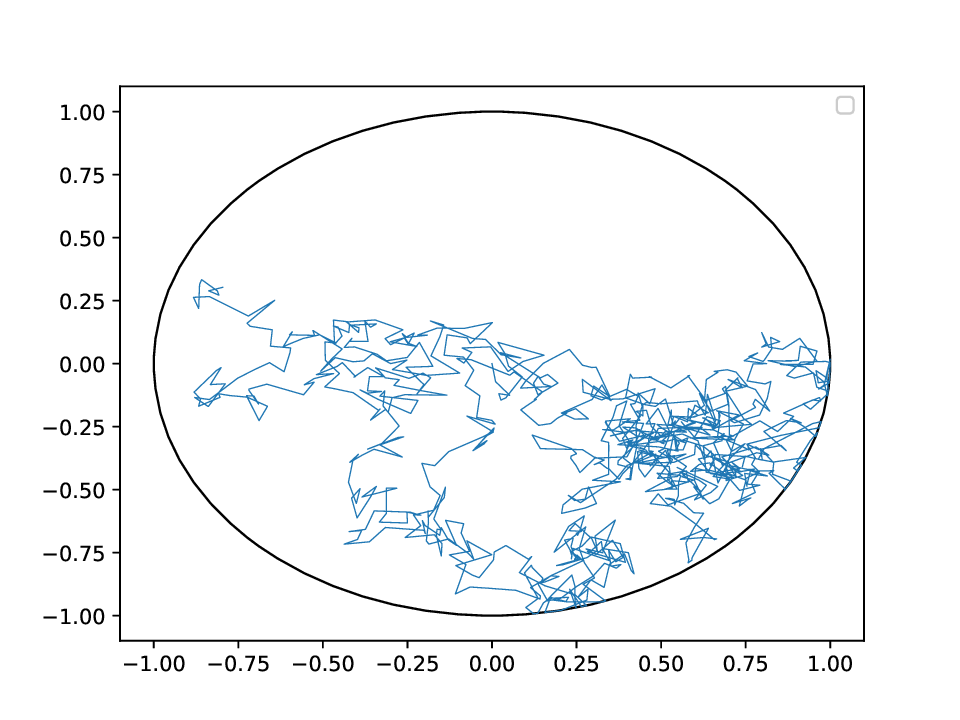}
        \caption{Trajectory of $X_t$ of reflected  (overdamped) Langevin dynamics. }
\end{figure}
\vspace{-4pt}
\begin{table}[htbp]
    \centering
    \caption{Comparison between confined Langevin dynamics (CLD) \eqref{cld_pos_for}-\eqref{cld_vel_for} and reflected SDEs \eqref{over_damp_Lang_sde}.}
  \label{comparison_table_1}
    \begin{tabularx}{\textwidth}{ cX X}  
         \toprule
        &\textbf{Confined Langevin dynamics (CLD)} & \textbf{Reflected (overdamped) Langevin dynamics} \\
         \midrule
      (i) &  Restriction to the constraints is achieved via a c\'{a}dl\'{a}g process which reflects the velocity if $X_t$ collides with the boundary &  Dynamics adheres to the constraints due to local time term. \\
      (ii) & Specular reflection is more natural as it ensures the angle of incidence and angle of reflection remains same.  & Reflection is along the pre-specified direction when $X_t \in \partial G$ showing that the boundary interferes in the dynamics.  \\
    
 (iii) &  $X_t$ is piecewise continuously differentiable function of $t$ almost surely. 
 & $X_t$ is continuous and non-differentiable function of 
 $t$ almost surely \\
    \bottomrule 
    \end{tabularx}
\end{table}

\section{Approximating Langevin diffusions with boundary condition}
In Section~\ref{sec_3.1_nrps_cld} and Section~\ref{sec_3.2_nrps_rsde}, we discuss the numerical implementation of  confined Langevin dynamics based models \eqref{cld_pos_for}-\eqref{cld_vel_for}  and reflected SDEs (\ref{over_damp_Lang_sde}), respectively, along with their corresponding sampling processes.
We partition $[0,T]$ as $0= t_0 <\cdots <  t_{N} = T$ such that $ t_{k+1} - t_{k} = h = T/N$. We take $\gamma$ and $b$ to be time-independent. Let $\xi_k = \{\xi^{1}_k, \dots, \xi^{d}_k \} $ with $\xi^{i}_k \sim \mathcal{N}(0,1)$, $i= 1,\dots, d$, $k=1,\dots,N$. 
\subsection{Simulating confined Langevin dynamics and its time reversal}\label{sec_3.1_nrps_cld}
For the purpose of simulating confined Langevin dynamics, we discuss two kinds of splittings here : [A, B, O] splitting \cite{leimkuhler2015molecular, bou2010long} and BBK splitting \cite{brunger1984stochastic} to implement (\ref{cld_pos_for})-(\ref{cld_vel_for}). The [A, B, O] has been extended to confined setting in \cite{leimkuhler2024numerical} which also forms the basis of numerical approximation of time reverse dynamics \eqref{rev_cld_1}-\eqref{rev_cld_2}. The extension of BBK scheme to confined setting is not considered earlier. 

The main concern in the numerical approximation is the collision with boundary. Let $(y,z)$ denote the current state of the Markov chain approximating (\ref{cld_pos_for})-(\ref{cld_vel_for}) or \eqref{rev_cld_1}-\eqref{rev_cld_2}. We are first discussing the collision procedure here so we denote the update after collision as $(Y_1, Z_1)$. 
Let $r$ be a binary valued number which takes value $0$ for forward dynamics else $1$. Let $\mathcal{Y}_1 := y + (-1)^{r}zh $.  As is clear, this corresponds to forward dynamics if $r = 0$ otherwise it is an update in reverse dynamics. If $\mathcal{Y}_1 \in \bar{G} $ then we assign $Y_1 = y + (-1)^{r} zh$ and $Z_1 = z$. However, if the chain collides with the boundary i.e. $\mathcal{Y}_1= y + (-1)^{r} z h \notin \bar{G}$, the procedure to get next state of chain is as follows: 
we find $\tau \in (0,h)$ such that $ \mathcal{Y}_{\tau}:= y + \tau (-1)^{r} z \in \partial G$. Then, the momentum is reverted at $\mathcal{Y}_{\tau}$ and an update in position follows using the reverted velocity $Z_1$ with remaining time $h- \tau$ as described in Figure~\ref{Collision_step}.
\begin{figure}[htbp]
\begin{equation*}
A_c = \left\{
    \begin{aligned}
    \mathcal{Y}_\tau  &= y + (-1)^{r}z \tau \in \partial G, \\
    {Z}_1 &= z - 2 \la n(\mathcal{Y}_\tau) \cdot z\ra n(\mathcal{Y}_\tau), \\
    {Y}_1 &=  \mathcal{Y}_\tau  + (-1)^{r}(h-\tau){Z}_1.  
    \end{aligned}
\right.
\end{equation*}
\caption{Collision step if $\mathcal{Y}_1 \notin \bar{G}$ i.e. if auxiliary step is outside domain $G$.}\label{Collision_step}
\vspace{0pt}
\end{figure}

In compact form, we write the collision step as
\begin{align*}
   ({X}_1, {V}_1) :&=  A_c(x,v, h),\; \text{in case of forward dynamics: }r=0, \\
   ({Q}_1, {P}_1) :&=  A_c(q,p, h),\; \text{in case of reverse dynamics: }r=1.
\end{align*}
\paragraph{Simplification in case of hypercube setting $[a_1, a_2]^d$.} The above collision procedure is significantly simplified in case of box constraints.  Let $\mathcal{Y}_{1}^{i} >a_2>0$ i.e. $i$-th component of $\mathcal{Y}_1^{i}$ cross the boundary, then the $A_c$ step is 
\begin{align}
    Y_{1}^{i} = 2a_2 I(\mathcal{Y}^{i}_1 > a_2 ) -
    \text{sgn}( \mathcal{Y}_1^i > a_2) \mathcal{Y}_1^i, 
\end{align}\vspace{0pt}
where $\text{sgn(TRUE)} = 1$ else $-1$. Similar expression can be obtained for $\mathcal{Y}_{1}^{i} <a_1$. 
 \vspace{-10pt}
\paragraph{Forward process.}

 \begin{figure}[htbp]
\vspace{-4pt}
\centering
   \begin{minipage}{.4\textwidth}
   \vspace{-4pt}
     \begin{flalign} &\textrm{\underline{[A$_c$OA$_c$]}} \nonumber \\   & (X_{k+1/2},V_{k+1/2}^{r}) =  A_c(X_k,V_{k};h/2)   \;\;\; \nonumber \\
      & \hat{V}_{k+1/2} =O(V_{k+1/2}^{r};h,\xi_{k+1},\gamma)   \nonumber\\
      &  (X_{k+1},V_{k+1}) =  A_c(X_{k+1/2},\hat{V}_{k+1/2};h/2).  \nonumber 
   \end{flalign}
   \end{minipage}
 \begin{minipage}{.4\textwidth}
\begin{flalign} &\textrm{\underline{[CBBK]}} \nonumber \\
   &   V_{k+ 1/2}  = V_k - \frac{h \gamma}{2} V_k + \sqrt{ \frac{\gamma h}{2}}\xi_k - \frac{h}{2}X_k \nonumber 
        \\  
      &  (X_{k+1}, \hat{V}_{k+1/2})   = A_c (X_k,  V_{k+1/2} , h) \nonumber \\
      &  V_{k+1} =  \frac{1}{1 + \frac{h\gamma}{2}}\big( \hat{V}_{k+1/2} + \frac{h}{2} X_{k+1} + \sqrt{\frac{\gamma h}{2}} \xi_{k+1}\big). \nonumber
   \end{flalign}
\end{minipage}
\caption{Description of [A$_c$OA$_c$] (for $b \equiv 0$) and [CBBK] (for $b = -x$).}\label{forward_aoa_cbbk}
\end{figure}
  As already mentioned that extension of [A, B, O] splitting schemes for CLD (\ref{cld_pos_for})-(\ref{cld_vel_for}) in forward setting is available in \cite{leimkuhler2024numerical}.  For the sake of completeness, we mention one of the possible splitting scheme when $ b \equiv 0$, i.e. [A$_c$OA$_c$] in Figure~\ref{forward_aoa_cbbk}. In this case, the dynamics is split into Orstein-Uhlenbeck dynamics (denoted as $O(v; h, \xi, \gamma):=  ve^{-\gamma h} + \sqrt{(1 - e^{-2 \gamma h})} \xi$) and collisional drift (denoted as A$_c$). Other possible symmetrized concatenation is [OA$_c$O]. In the similar manner, one can modify  BBK scheme \cite{brunger1984stochastic} by replacing positional update with A$_c$
 step (see Figure~\ref{forward_aoa_cbbk}) to obtained confined BBK scheme. 
\paragraph{Reverse process.}
We have already discussed the A$_c$ step. The $O$ step in sampling process is given by
$    O(p;  h , \xi, \gamma):=  pe^{\gamma h} + \sqrt{(e^{2 \gamma h} - 1)}\xi.
$ 
Given $(q,p)$, in addition to $O$ and  A$_c$, we also have the following update:
\begin{align}
      S(t,q,p; \gamma, h) := p - b(q) h + 2 \gamma s_{\theta}(T-t, q, p)h. 
\end{align}

We mention one symmetrized concatenation [SA$_c$OA$_c$S] in Figure~\ref{saoas_bbks}. 
It is not difficult to write other possible symmetrized concatenations namely :  [OSA$_c$SO], [A$_c$SOSA$_c$], [A$_c$OSOA$_c$] etc. 
For the reverse, we present the confined BBK (Brünger-Brooks-Karplus) scheme with score function in Figure~\ref{saoas_bbks}, denoted as CBBK-S, where the positional update is constrained to be in $\bar{G}$ by specular reflection of velocity via
 $A_{c}$ step as described in Figure~\ref{Collision_step}. 

\begin{figure}[ht]
\vspace{-4pt}
\centering
   \begin{minipage}{.4\textwidth}
   \vspace{-4pt}
\begin{flalign} &\textrm{\underline{[SA$_c$OA$_c$S]}} \nonumber \\
   & P_{k+1/2} =   S(t_k, Q_k, P_k; \gamma, h/2) \nonumber \\
    & (Q_{k+1/2}, P_{k+1/2}^{r}) =  A_c(Q_k,P_{k+1/2};h/2)   \;\;\; \nonumber \\
      & \hat{P}_{k+1/2} =O(P_{k+1/2}^{r};h,\xi_{k+1},\gamma)   \nonumber\\
      &  (Q_{k+1},P_{k+1}^{r}) =  A_c(Q_{k+1/2},\hat{P}_{k+1/2};h/2)  \nonumber \\
    &P_{k+1} = S(t_k, Q_{k+1}, P_{k+1}^{r}; \gamma, h/2), \nonumber
   \end{flalign}
\end{minipage}
 \begin{minipage}{.4\textwidth}
\begin{flalign} &\textrm{\underline{[CBBK-S]}} \nonumber \\
   &   P_{k+ 1/2}  = P_k + \frac{h \gamma}{2} P_k + \sqrt{ \frac{\gamma h}{2}}\xi_k + \frac{h}{2}Q_k \nonumber 
        \\  
      &  (Q_{k+1}, \hat{P}_{k+1/2})   = A_c (Q_k,  P_{k+1/2} , h) \nonumber \\
      &  P_{k+1} =  \frac{1}{1 - \frac{h\gamma}{2}}\big( \hat{P}_{k+1/2} + \frac{h}{2} Q_{k+1} + \sqrt{\frac{\gamma h}{2}} \xi_{k+1} \nonumber \\  &  \quad \quad + 2h \gamma s_{\theta}(T-t_k, Q_{k+1}, \hat{P}_{k+ 1/2})\big).  \nonumber 
   \end{flalign}
\end{minipage}
\caption{Description of [SA$_c$OA$_c$S] and [CBBK-S].}\label{saoas_bbks}
\end{figure}
As is clear from the description of [SA$_c$OA$_c$S] and [OSA$_c$SO], these splitting schemes require two evaluations of score network. The two evaluation of score network per iteration may be detrimental for the performance for large networks.  To this end, we also propose different schemes  namely 
[BA$_c$OA$_c$S] and [OBA$_c$SO],
where B step corresponds to update in $P$ but using only $b$ i.e.
\begin{align}
    B(q, p;h) = p - b(q) h. 
\end{align}
The description for [BA$_c$OA$_c$S] and [OBA$_c$SO] is in Figure~\ref{baoas_desc_appen}. It is not difficult to write other possible concatenations, for example [SA$_c$OA$_c$B] or [A$_c$SOBA$_c$] etc. For the sake of avoiding any confusion, we write 1FE [SA$_c$OA$_c$S] instead of [SA$_c$OA$_c$B] to highlight it is the same scheme. We will explicitly mention if we are using 1FE or 2FE [SA$_c$OA$_c$S] wherever required.

\begin{figure}[H]
\centering
   \begin{minipage}{.4\textwidth}
\begin{flalign} &\textrm{\underline{[BA$_c$OA$_c$S]}} \nonumber \\
   & P_{k+1/2} =   B( Q_k, P_k; h/2) \nonumber \\
    & (Q_{k+1/2}, P_{k+1/2}^{r}) =  A_c(Q_k,P_{k+1/2};h/2)   \;\;\; \nonumber \\
      & \hat{P}_{k+1/2} =O(P_{k+1/2}^{r};h,\xi_{k+1},\gamma)   \nonumber\\
      &  (Q_{k+1},P_{k+1}^{r}) =  A_c(Q_{k+1/2},\hat{P}_{k+1/2};h/2)  \nonumber \\
    &P_{k+1} = S(t_k, Q_{k+1}, P_{k+1}^{r}; 2\gamma, h/2), \nonumber
   \end{flalign}
\end{minipage}
 \begin{minipage}{.4\textwidth}
\begin{flalign} &\textrm{\underline{[OBA$_c$SO]}} \nonumber \\
   & P_{k+1/2} =O(P_{k};h/2,\xi_{k+1},\gamma)  \nonumber \\
   & \hat{P}_{k+1/2} = B(t_k,Q_k, P_{k+1/2}; h/2) \nonumber\\
    & (Q_{k+1}, P_{k+1/2}^{r}) =  A_c(Q_k,\hat{P}_{k+1/2};h)   \;\;\; \nonumber \\
      & \hat{P}_{k+1} =S(Q_{k+1}, P_{k+1/2}^{r};2\gamma, h/2)   \nonumber\\
      &  P_{k+1} =  O( \hat{P}_{k+1}; h/2, \gamma).  \nonumber 
   \end{flalign}
\end{minipage}
\caption{Description of [BA$_c$OA$_c$S] and [OBA$_c$SO]} \label{baoas_desc_appen}
\end{figure}

\subsection{Simulating reflected (overdamped) Langevin SDE and its time reversal}\label{sec_3.2_nrps_rsde}
We define the projection operator for $x \in \bar{G}^c$ and Eulerian update (given $X_k$) as follows:
\begin{align}
    \Pi(x) &= \arg \min_{y \in \partial G} |x- y|^2 \quad\text{with assignment}\quad   \Pi(x) = x, \; x \in \bar{G}, \\  
  \delta_{k+1} &= f(t_k, X_{k})h  + \sqrt{2}\xi_{k+1}\sqrt{h},\label{aux_step} \quad \quad\quad 
\end{align}
where $\xi_{k+1}^{i} \sim \mathcal{N}(0,1)$, $i= 1,\dots, d$ and $k= 0,\dots, N-1$. For forward dynamics $f(t, x) := b(x)$ and for reverse/sampling process $f(t , x) := -b(x) + 2s_{\theta}(T-t, x)$. 

All the method mentioned below (except barrier method) have been considered for approximating reflected SDEs. The reflected diffusion model from \cite{lou2023reflected} proposed to use projection method.

\paragraph{(i) Projection method.}
We denote an auxiliary step by
\begin{align}\label{cld_gen_eq_15}
    X^{'}_{k+1} = X_k + \delta_{k+1},
\end{align}
where $\delta_{k+1}$ is from \eqref{aux_step}. 
If $X^{'}_{k+1} \in \bar{G}$, then $X_{k+1} = X^{'}_{k+1}$, else $X_{k+1} = \Pi(X^{'}_{k+1})$ (see \cite{constantini_proj} for half-order convergence result in weak-sense).  

    \paragraph{(ii) Symmetrized reflected method.}  Consider the auxiliary step in (\ref{cld_gen_eq_15}). Symmetrized reflection scheme \cite{sym_euler_bossy_gobet_talay, leimkuhler2023simplest} which is a first-order method (in weak-sense) without additional cost, is given by 
\begin{align}\label{sym_ref_Scheme}
    X_{k+1} = \begin{cases}
        X_{k+1}^{'}, \quad \text{if}\quad   X_{k+1}^{'} \in \bar{G},  \\ 
        X_{k+1}^{'} - 2 d_{k+1} n(\Pi(X_{k+1}^{'})),  \quad &\text{else, where}  \quad
 d_{k+1} := |X_{k+1}^{'} - \Pi(X_{k+1}^{'})|.
    \end{cases}
\end{align} 
Note that $X_{k+1}^{'} - d_{k+1} n(\Pi(X_{k+1}^{'}))$ represents the projection of $X_{k+1}^{'}$ on $\partial G$. 

\paragraph{(iii) Penalty method}
This method penalizes the Markov chain as soon as it leaves the constrained set $\bar{G}$. The one-step procedure is given as
\begin{align}
    X_{k+1} &= X_k + \delta_{k+1} - \frac{h}{\lambda}(X_{k} - \Pi(X_{k})), \label{penalty_method}
\end{align}
where $\lambda>0$ denotes a penalty parameter. The convergence, in strong sense, has been shown in \cite{slominski2001euler}.  
\paragraph{(iv) Barrier method} Consider a neighborhood of $\partial G$ defined as $G_{\epsilon} : = \{ x \in  G \; ; \; \dist(x, \partial G) \leq \epsilon   \}$ for some $\epsilon > 0$. Also, consider function $R(x)$ which is equal to $\dist(x, \partial G)$ for $x \in G_{\epsilon}$ else it is greater than some constant $\epsilon_0$. Inspired from convergence of barrier SDEs to reflected SDEs in \cite{arnaudon_xue_mei2017reflected}, we present here the barrier method:
\begin{align}
    X_{k+1} &= X_k + \delta_{k+1} + h\nabla \ln\bigg( \frac{\tanh{R(X_k)}}{\eta}\bigg), \; \text{with}\;  \nabla \ln\bigg( \frac{\tanh{R(x)}}{\eta}\bigg)  = \frac{2 \nabla R(x)}{\eta \sinh \frac{2R(x)}{\eta}},\label{barrier_method}
\end{align}
where, for $x \in G_{\epsilon} $,  $R(x)$ is $\dist(x, \partial G)$ 
and $
    \nabla R(x) = \frac{x - \Pi(x)}{|x - \Pi(x)|}.$ 
Here, $\eta >0$ denotes the barrier parameter. 
Here are some important observations:
\begin{itemize}
    \item The penalty method (\ref{penalty_method}) penalizes when the Markov chain exits the constraint set $G$, while the barrier method \eqref{barrier_method} creates a large barrier as the chain approaches the boundary $\partial G$ from within.
    \item Unlike the barrier method in \cite{fishman2023diffusion}, which transforms
$G$ into a Riemannian manifold by inducing a metric under which  the boundary is at infinite distance, we treat $G$ as a submanifold with boundary under the Euclidean metric, avoiding the need for exponential map-based schemes (see \eqref{barrier_method}).   
\end{itemize}

\section{Score-matching loss functions}
We here discuss the two score matching loss functions corresponding to confined Langevin diffusion model and reflected diffusion model. 
\begin{proposition}\label{prop_cld}
    Let $s_{\theta} \in C^{1}([0,T] \times \mathbb{R}^{d} ; \mathbb{R}^{d})$. Then, the score matching loss function for CLD (\ref{cld_pos_for})-(\ref{cld_vel_for}) is given by:
    \begin{align}
    \mathbb{E}_{t\in \mathcal{U}([0,T])}\mathbb{E}_{\mathcal{L}_{t}^{(X,V)}}\big(| s_{\theta}(t, X_t, V_t)|^2 + 2\diver_{v}(s_{\theta}(t, X_t, V_t)) \big) + C,
\end{align}
where $C >0$ is a constant and $\mathcal{L}_{t}^{(X,V)}$ is the time-marginal of joint law of $(X,V)$ from (\ref{cld_pos_for})-(\ref{cld_vel_for}). 
\end{proposition}
The loss function above is along the lines of \cite{hyvarinen2005estimation, dockhorn2022scorebased}. 

\begin{proposition}\label{ref_lossProp_2}
Let $s_{\theta} \in C^{1}([0,T] \times \mathbb{R}^{d} ; \mathbb{R}^{d})$. Then, the score matching loss function for reflected dynamics (\ref{over_damp_Lang_sde}) is given by:
\begin{align}
    \mathbb{E}_{t \sim \mathcal{U}([0,T])}\mathbb{E}_{\mathcal{L}_{t}^{X}}\bigg(|s_{\theta}(t, X_t)|^2  + 2 \diver (s_{\theta}(t, X_t))  
  - \frac{2}{t}\int_{0}^{t}\langle s_\theta(r, X_r) \cdot n(X_r)\rangle  \de L_r\bigg) + C, \label{prop_2_eq_27}
\end{align}    
where $C >0$ is a constant, $\mathcal{L}_{t}^{X}$ is time-marginal law of reflected dynamics (\ref{over_damp_Lang_sde}) and $L_r$ is local time from   (\ref{over_damp_Lang_sde}). 
\end{proposition}
It is clear that we can use Monte-Carlo estimator for loss function in Proposition~\ref{prop_cld} and for first two terms in \eqref{prop_2_eq_27}. In next proposition, we demonstrate that the third term in \eqref{prop_2_eq_27} can, in fact, be effectively computed.  
\begin{proposition}\label{ref_lossProp_3}
Let $\partial G $ be $C^4$ and $b \in C^2( \bar{G})$ in \eqref{over_damp_Lang_sde}. Then, we have
\begin{align}
    \bigg| \mathbb{E}\frac{1}{t}\int_{0}^{t}\langle s_\theta(r, X_r) \cdot n(X_r)\rangle  \de L_r 
     - \frac{2}{t} \mathbb{E}\sum_{k=0}^{\lfloor t/h \rfloor -1 }d_{k+1}\la s_{\theta}' \cdot n_{k+1}^{\pi}\ra I_{G^c}(X_{k+1}')\bigg| \leq  C h,
\end{align}
where $s_{\theta}' := s_{\theta}(t_k, X_{k+1}')$, $n_{k+1}^{\pi} :=n(\Pi(X_{k+1}'))$, and $X_{k+1}$, $X_{k+1}'$ and $d_{k+1}$ are from (\ref{sym_ref_Scheme}) and $C >0$ is a generic constant independent of $h$. 
\end{proposition}
For the projection scheme, the estimator remains the same except the $2$ in the numerator.  The proofs are provided in Appendix~\textcolor{red}{A}.

\begin{figure}[H]
    \centering
    \begin{subfigure}{0.45\textwidth}
        \includegraphics[width=0.9\linewidth]{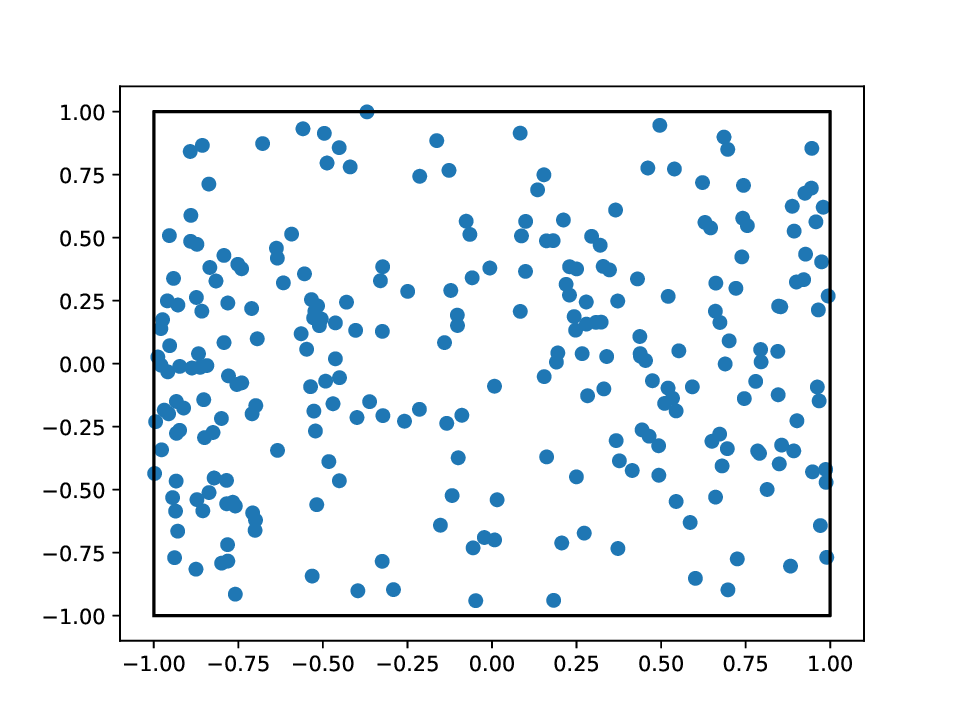}
        \caption{Gaussian mixture $[-1,1]$ with ignoring non-tractable term in loss}
        \label{fig:gaussian_mixture_wo_extraterm}
    \end{subfigure}%
    \hfill 
    \begin{subfigure}{0.45\textwidth}
        \includegraphics[width=0.9\linewidth]{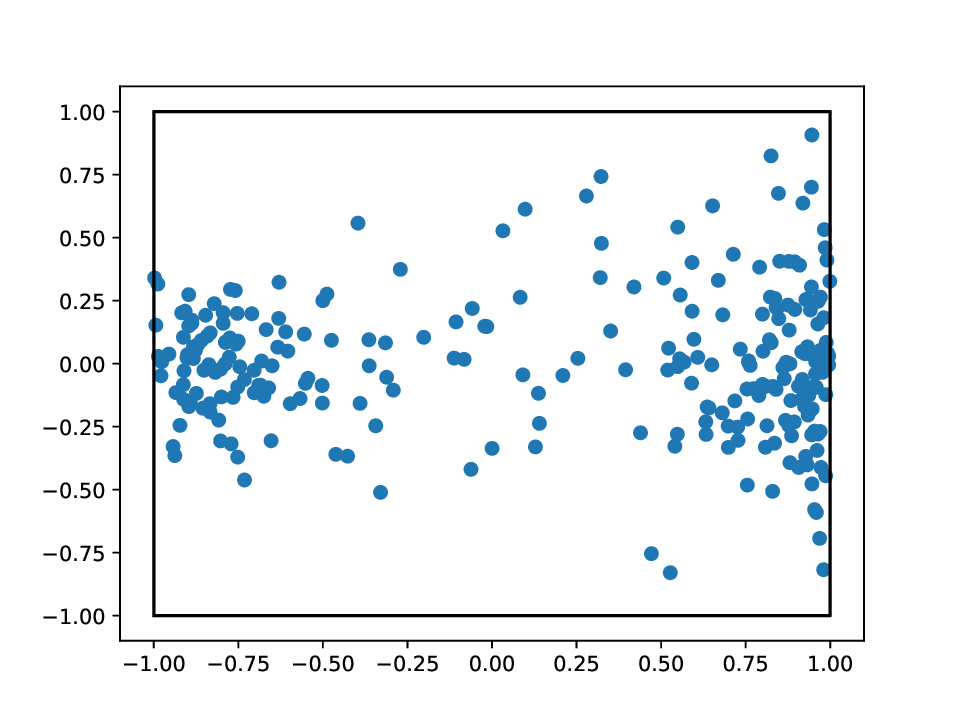}
        \caption{Gaussian mixture $[-1,1]$ with the proposed loss function in Propositions \ref{ref_lossProp_2}-\ref{ref_lossProp_3}}
        \label{fig:gaussian_mixture_w_extraterm}
    \end{subfigure}
    \caption{Comparing loss functions for symmetrized reflection model}
    \label{fig:2combined_gaussian_mixture}
\end{figure}

\paragraph{Effect of corrected loss function}
We noticed that the effect of loss function with local time correction (see Propositions~\ref{ref_lossProp_2}-\ref{ref_lossProp_3}) is far more pronounced if the size of $G$ is small . We confirmed this with $G = (0,1)^2$ for two blobs produced using 2 Gaussian mixtures on the side of square and see Figure~\ref{fig:2combined_gaussian_mixture} for the generated samples. In this small experiment we have taken $h = 0.01$ and $T = 1$ and neural network is trained for 5000 iterations.  

\section{Experiments} \label{sec_5}

 In this section, we illustrate the performance of generative models introduced in earlier sections on toy datasets and a subset of MNIST dataset.  The models for the toy datasets are trained using an NVIDIA GeForce RTX 3060 mobile 6GB VRAM using a four layered MLP with hidden layers of size 128. Neural Net for MNIST  dataset is trained on an NVIDIA A40 48GB VRAM.   In Section~\ref{sec_5.1},  we compare reflected diffusion models (with forward dynamics \eqref{subsec_2.2_genrsde},  reverse dynamics \eqref{time_reverse_dyn} and the corresponding samplers from Section~\ref{sec_3.2_nrps_rsde}) and confined Langevin diffusion model (with forward dynamics \eqref{cld_pos_for}-\eqref{cld_vel_for}, reverse dynamics \eqref{rev_cld_1}-\eqref{rev_cld_2} and the corresponding samplers from Section~\ref{sec_3.1_nrps_cld}) for four toy data sets. In Section~\ref{sec_5.3}, we consider MNIST dataset.

\subsection{Comparison of models} \label{sec_5.1}
\paragraph{Gaussian mixture.} The dataset is generated from four normal distributions (variance $0.2^4$) with means on the sides of boundaries, excluding points outside the box $\bar{G}=\left[-3,3\right]^{2}$. The aim is to compare the several constrained diffusion generative models corresponding to CLD (\ref{cld_pos_for})-(\ref{cld_vel_for}), reflected SDEs (\ref{over_damp_Lang_sde}),  together with the DDPM \cite{ho2020denoising}. In all these models $b = -x$ (and $\gamma = 1$ for CLD models).  All models are trained for $5000$ iterations using the whole dataset (i.e. without batching) for $T=1$ and $h=0.005$.  
\begin{figure}[H]
\vspace{-4pt}
    \centering
    \begin{minipage}{0.7\linewidth}
        \centering
        \begin{subfigure}{0.24\linewidth}
            \centering
            \includegraphics[width=0.99\linewidth]{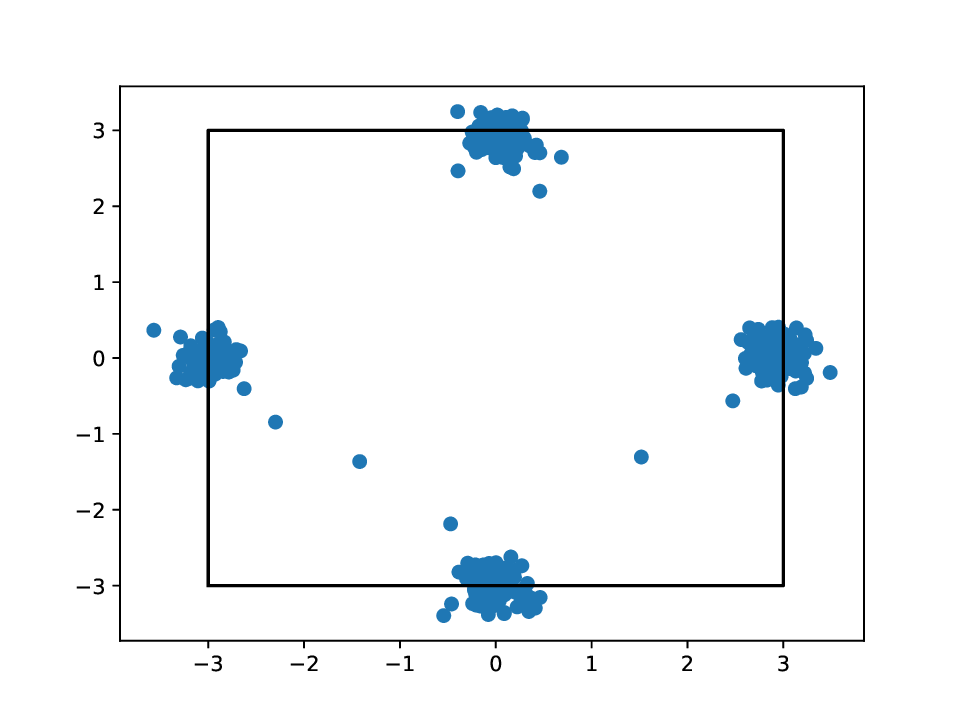}
            \caption*{DDPM}
            \label{fig:ddpm_left}
        \end{subfigure}
        \hfill
        \begin{subfigure}{0.24\linewidth}
            \centering
            \includegraphics[width=0.99\linewidth]{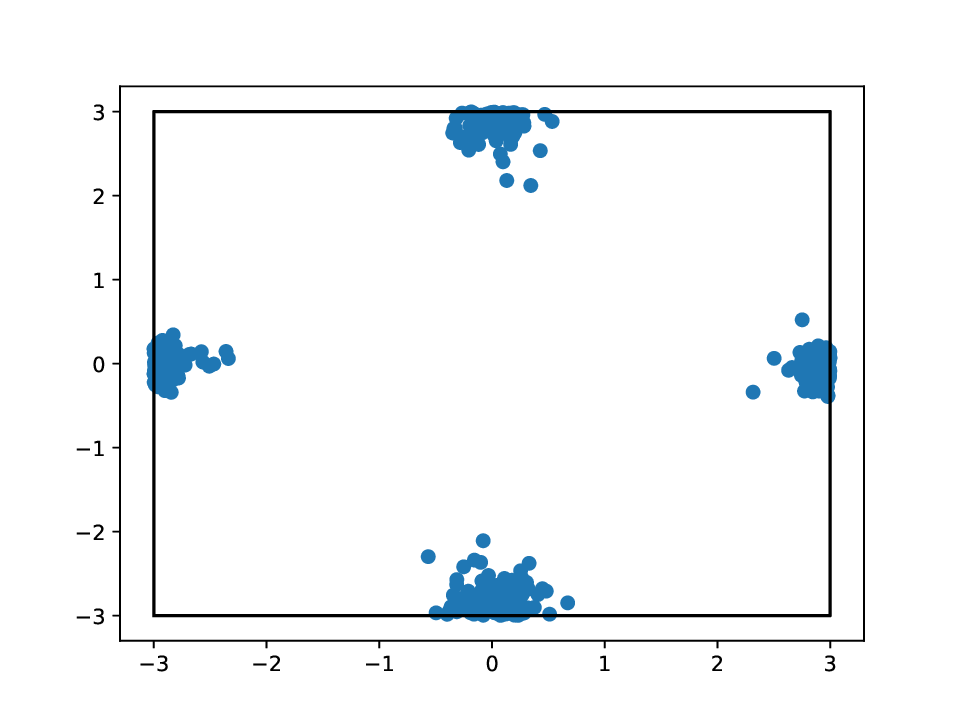}
            \caption*{Reflection}
            \label{fig:reflection_left}
        \end{subfigure}%
        \hfill
        \begin{subfigure}{0.24\linewidth}
            \centering
            \includegraphics[width=0.99\linewidth]{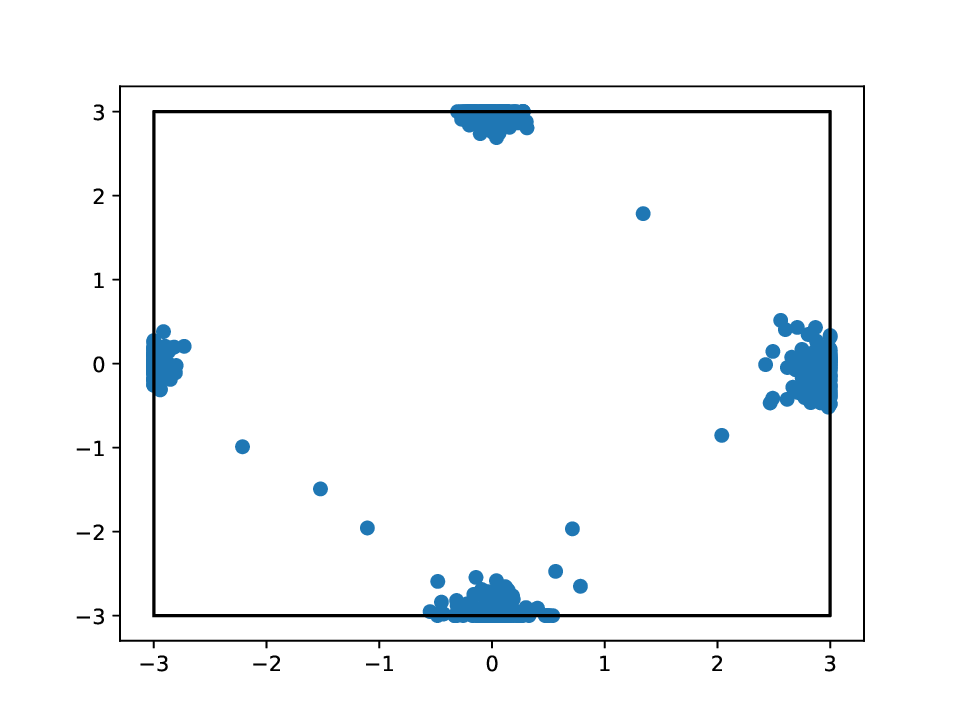}
            \caption*{Projection}
            \label{fig:projection_left}
        \end{subfigure}%
        \hfill
         \begin{subfigure}{0.24\linewidth}
            \centering
            \includegraphics[width=0.95\linewidth]{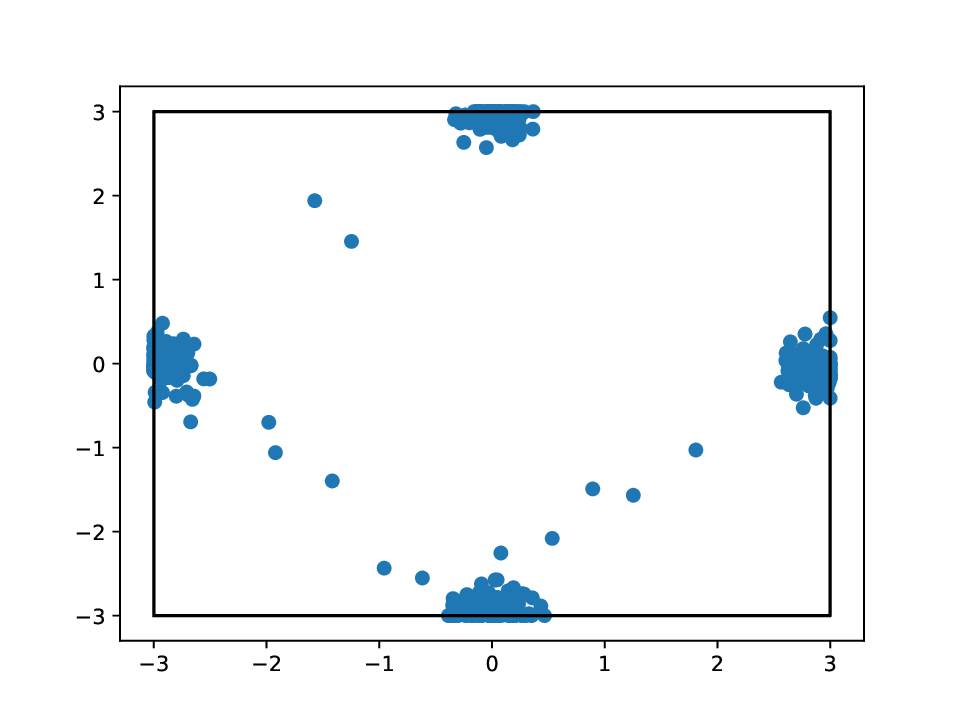}
            \caption*{Penalty}
            \label{fig:penalty_left}
        \end{subfigure}%
 \hfill
         \vspace{0.5em}

        \begin{subfigure}{0.24\linewidth}
            \centering
            \includegraphics[width=0.99\linewidth]{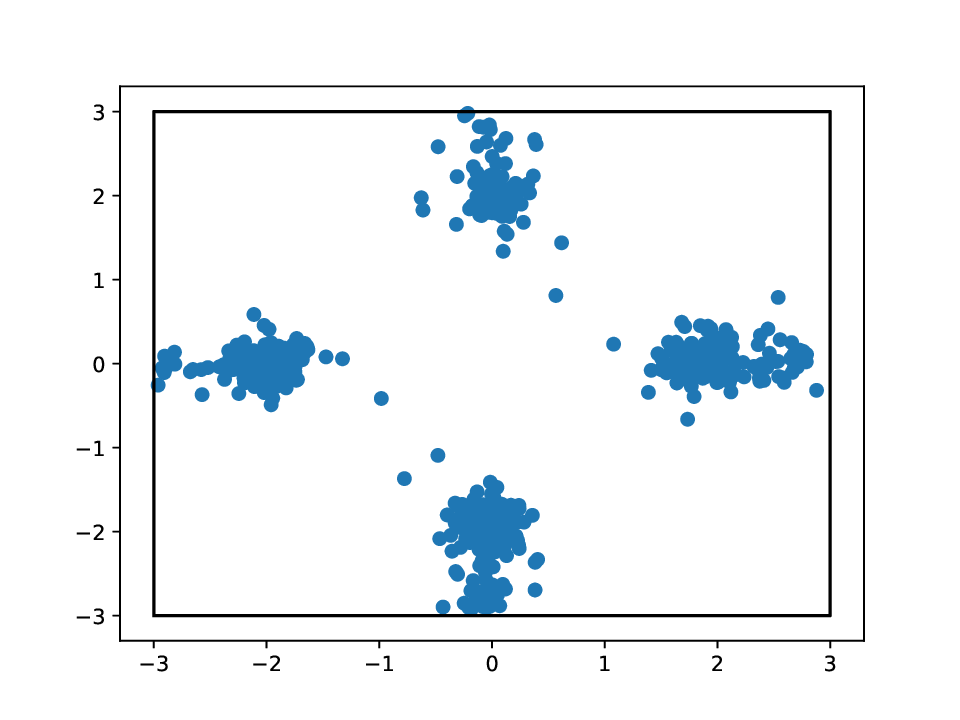}
            \caption*{Barrier}
            \label{fig:barrier_left}
        \end{subfigure}
         \hfill
        \begin{subfigure}{0.24\linewidth}
            \centering
            \includegraphics[width=0.99\linewidth]{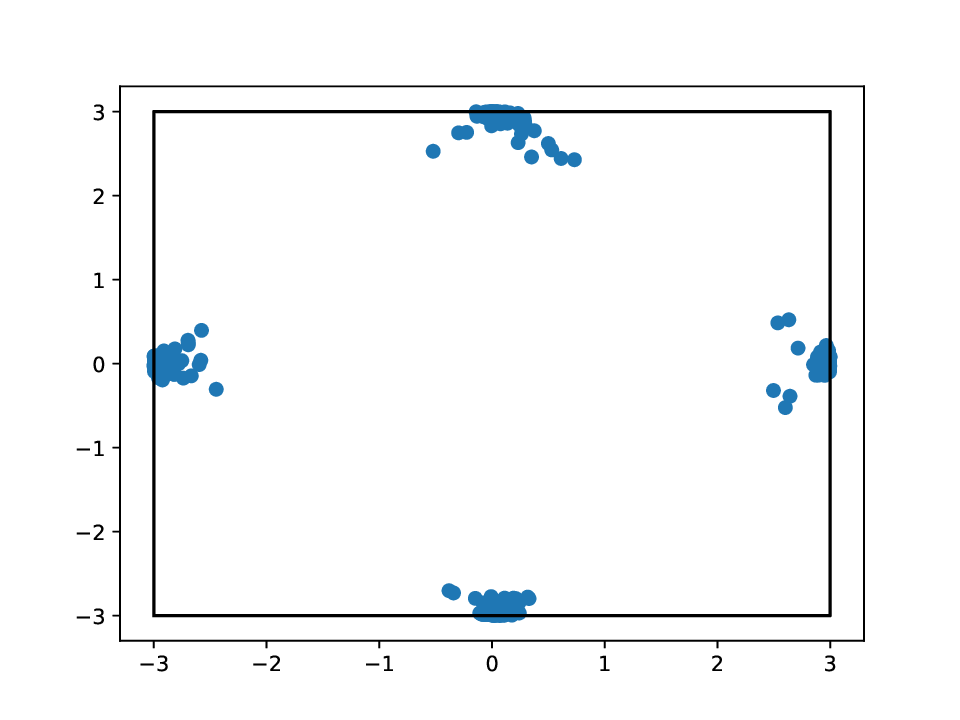}
            \caption*{SA$_c$OA$_c$S}
            \label{fig:baoab_left}
        \end{subfigure}%
         \hfill
        \begin{subfigure}{0.24\linewidth}
            \centering
            \includegraphics[width=0.99\linewidth]{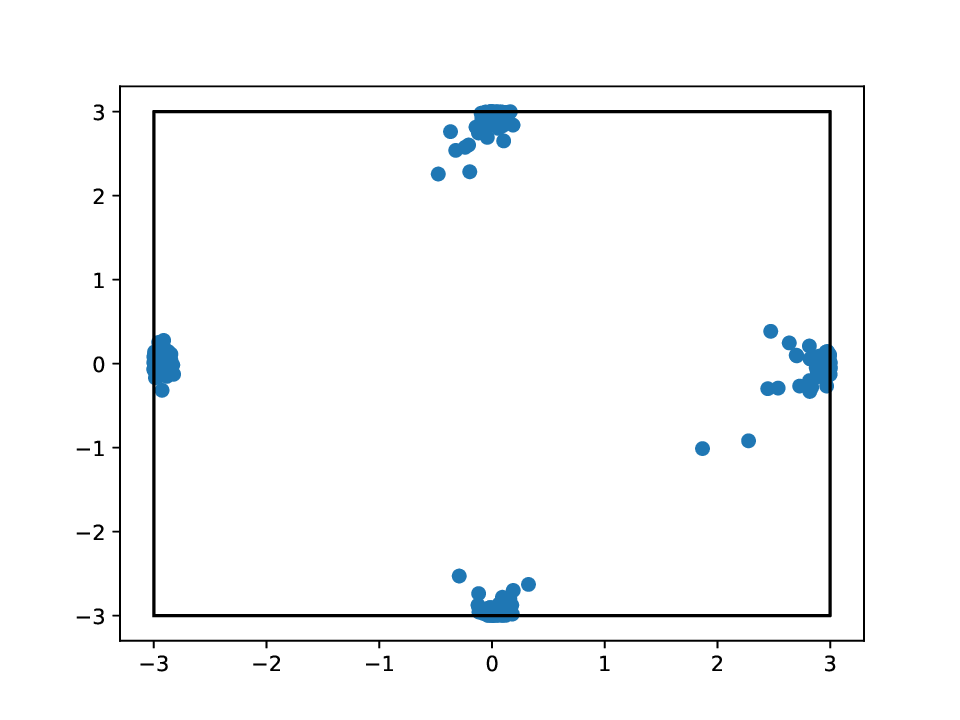}
            \caption*{OSA$_c$SO}
            \label{fig:obabo_left}
        \end{subfigure}%
        \hfill
        \begin{subfigure}{0.24\linewidth}
            \centering
            \includegraphics[width=0.95\linewidth]{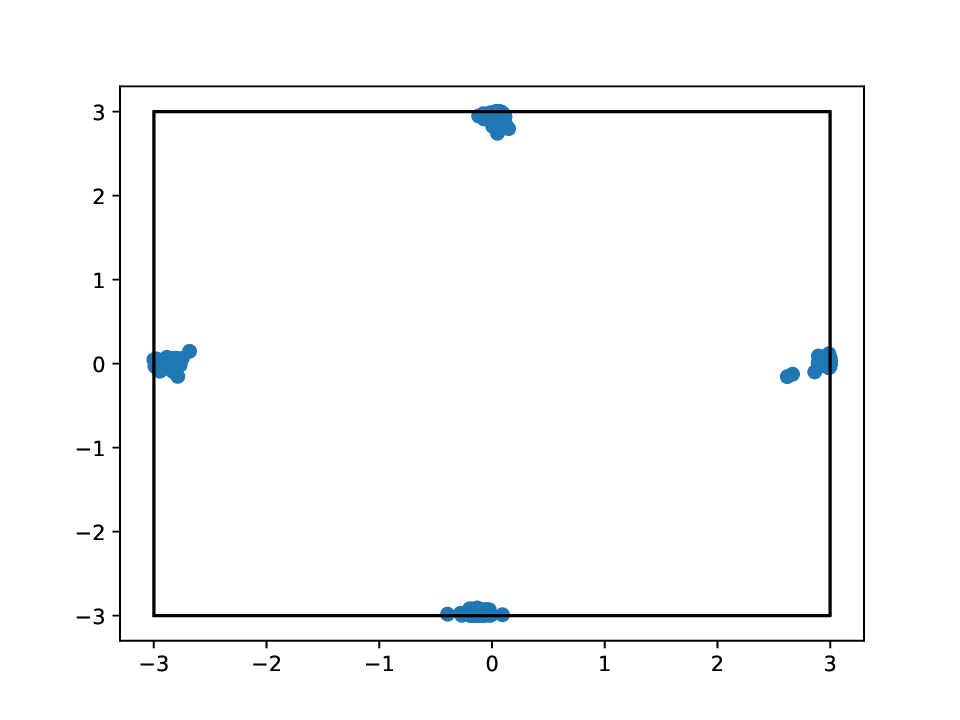}
            \caption*{CBBK-S}
            \label{fig:cbbk_left}
        \end{subfigure}%
        \hfill
        \caption{Generated Gaussian mixture samples}
        \label{fig:gm_sample_plots_left}
    \end{minipage}%
    \hfill
    \begin{minipage}{0.4\linewidth}
        \centering
        \includegraphics[width=0.99\linewidth]{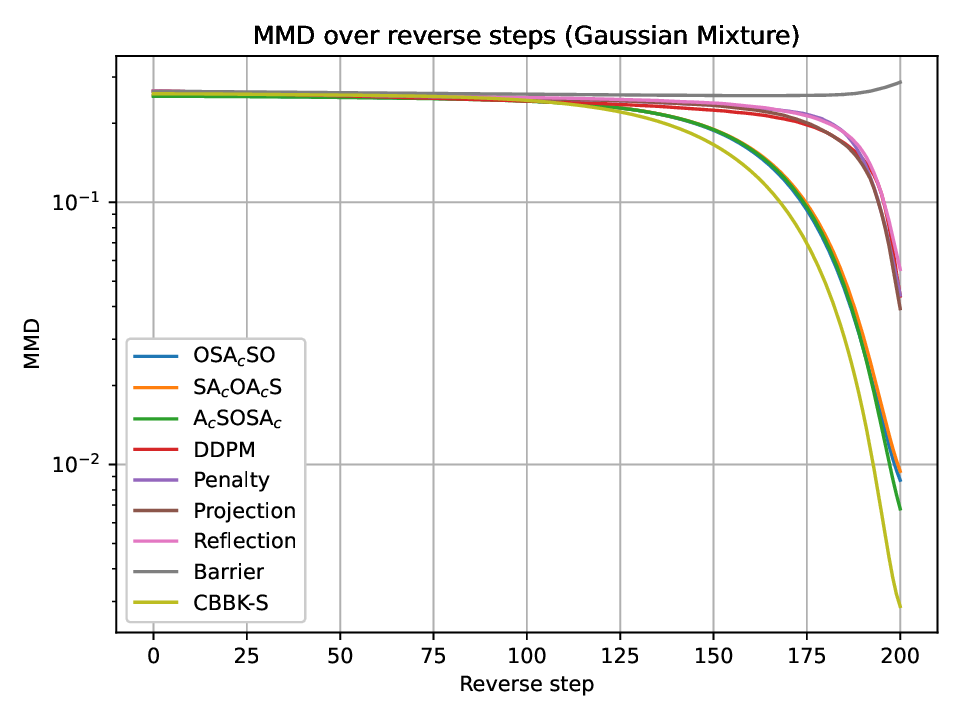}
        \caption{Model comparison using the Gaussian mixture dataset.}
        \label{fig:gm_mmd_comparison_right}
    \end{minipage}
\end{figure}

\begin{figure}[H]
    \centering
    \begin{subfigure}{0.2\linewidth}
        \centering
        \includegraphics[height = 1.7cm, width=0.90\linewidth]{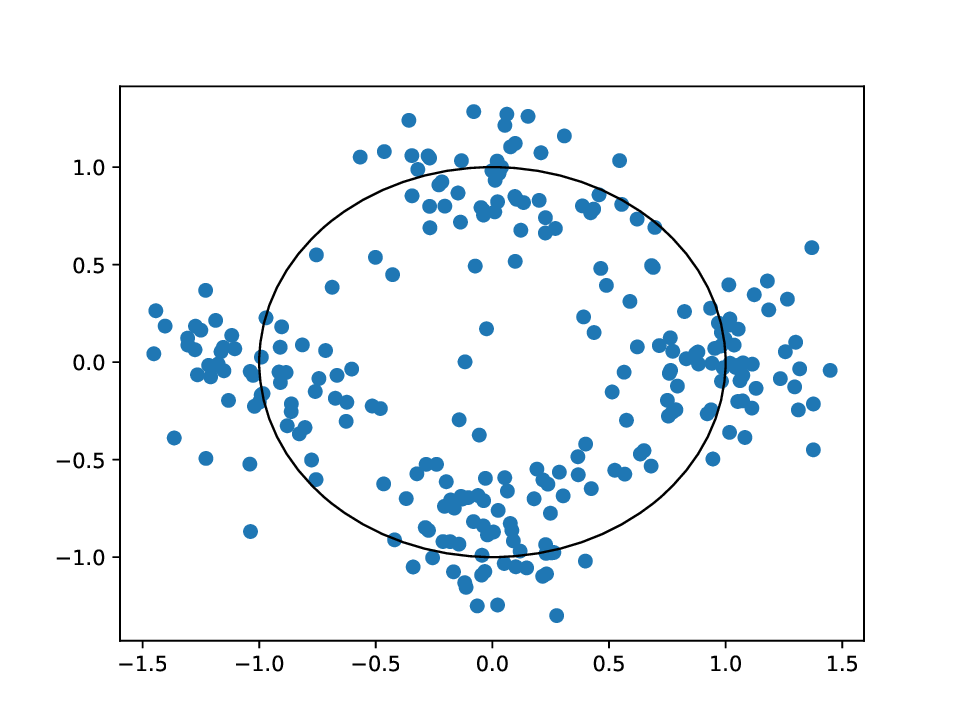}
        \caption*{DDPM}
        \label{fig:ddpm}
    \end{subfigure}
    \hfill
    \begin{subfigure}{0.2\linewidth}
        \centering
        \includegraphics[height = 1.7cm, width=0.90\linewidth]{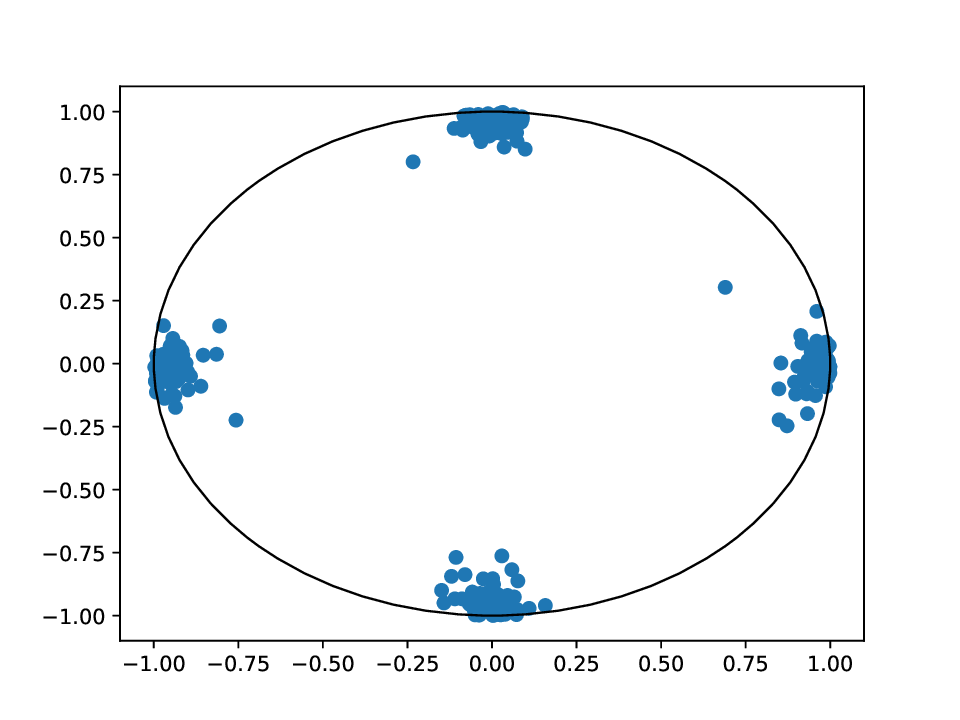}
        \caption*{SA$_c$OA$_c$S}
        \label{fig:baoab}
    \end{subfigure}
    \hfill
    \begin{subfigure}{0.2\linewidth}
        \centering
        \includegraphics[height = 1.7cm, width=0.90\linewidth]{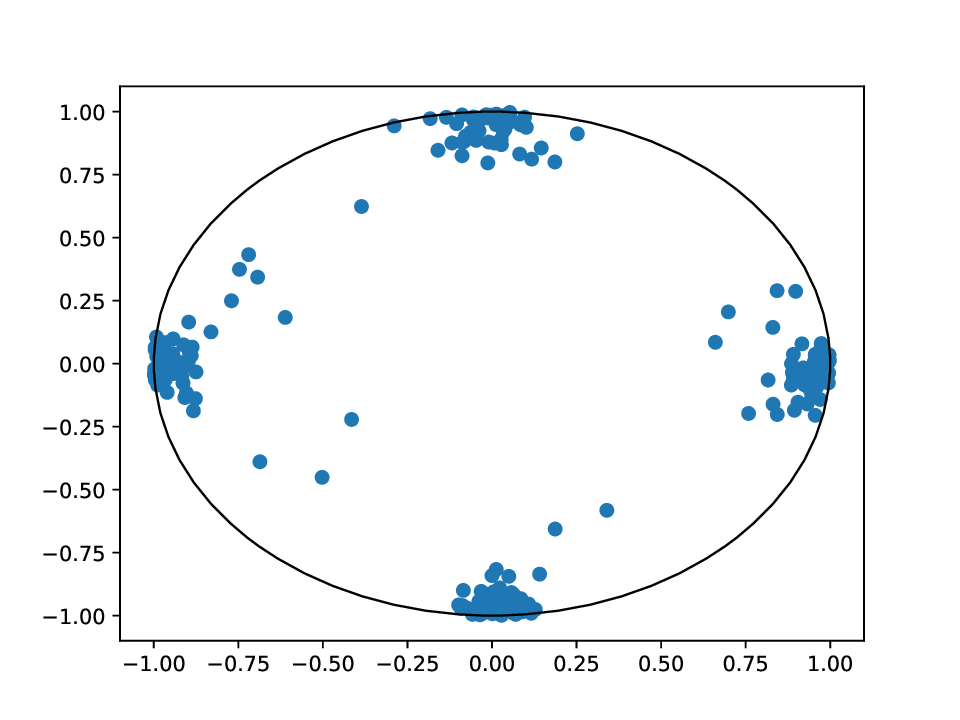}
        \caption*{OSA$_c$SO}
        \label{fig:osaso}
    \end{subfigure}
    \hfill
     \begin{subfigure}{0.2\linewidth}
        \centering
        \includegraphics[height = 1.7cm, width=0.90\linewidth]{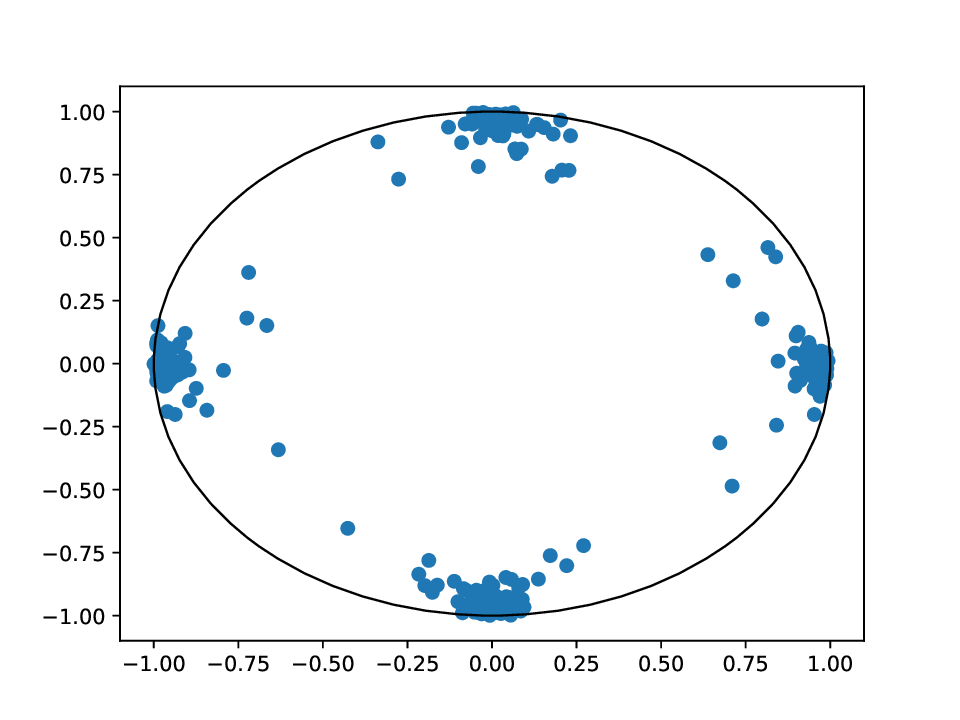}
        \caption*{A$_c$SOSA$_c$}
        \label{fig:asosa}
    \end{subfigure}
    \caption{Comparison with circular constraints} \label{circ_constr}
    \vspace{-10pt}
\end{figure}

The Maximum Mean Discrepancy (MMD) \cite{JMLR:v13:gretton12a} is calculated at each iteration of the sampling process. As is clear in Figure~\ref{fig:gm_mmd_comparison_right}, the CBBK-S resulted in the lowest MMD followed by [S, A$_c$, O] splitting schemes. The percentage of generated samples which violated box constraints are $37.41 \pm 0.41\%$  and $3.81 \pm 0.16 \%$ for DDPM and penalty method, respectively, and $0\%$ for all other models (see Table~\ref{tab:gm_mmd_comparison}). We obtain the similar results for circular constraints as can be seen in Figure~\ref{circ_constr}.

\begin{table}[h!]
\centering
\caption{Evaluation Metrics for Gaussian Mixture Experiment}
\label{tab:gm_mmd_comparison}
\begin{tabular}{lcc}
\toprule
Model & MMD ($\times 10^{-2}$) & Constraint Violation ($\%$) \\
\midrule
OSA$_c$SO & $0.89 \pm 0.02$ & $0.00 \pm 0.00$ \\
SA$_c$OA$_c$S & $0.94 \pm 0.03$ & $0.00 \pm 0.00$ \\
A$_c$SOSA$_c$ & $0.66 \pm 0.02$ & $0.00 \pm 0.00$ \\
CBBK-S & $0.28 \pm 0.01$ & $0.00 \pm 0.00$ \\

Penalty & $15.98 \pm 0.08$ & $3.81 \pm 0.16$ \\
Projection & $16.47 \pm 0.08$ & $0.00 \pm 0.00$ \\
Reflection & $16.23 \pm 0.09$ & $0.00 \pm 0.00$ \\
Barrier & $8.62 \pm 0.08$ & $0.00 \pm 0.00$ \\
DDPM & $14.53 \pm 0.07$ & $37.41 \pm 0.41$ \\
\bottomrule
\end{tabular}
\end{table}

\textbf{Wheel dataset.} 
This dataset contains $1232$ points lying inside $\left[-3,3\right]^2$.  We again test all schemes corresponding to reflected diffusion models and confined Langevin models, and present our results in Figure~\ref{fig:nfe_wheel_SAOAS_2-120}.

\begin{figure}[H]
    \centering
    \includegraphics[width=0.7\linewidth]{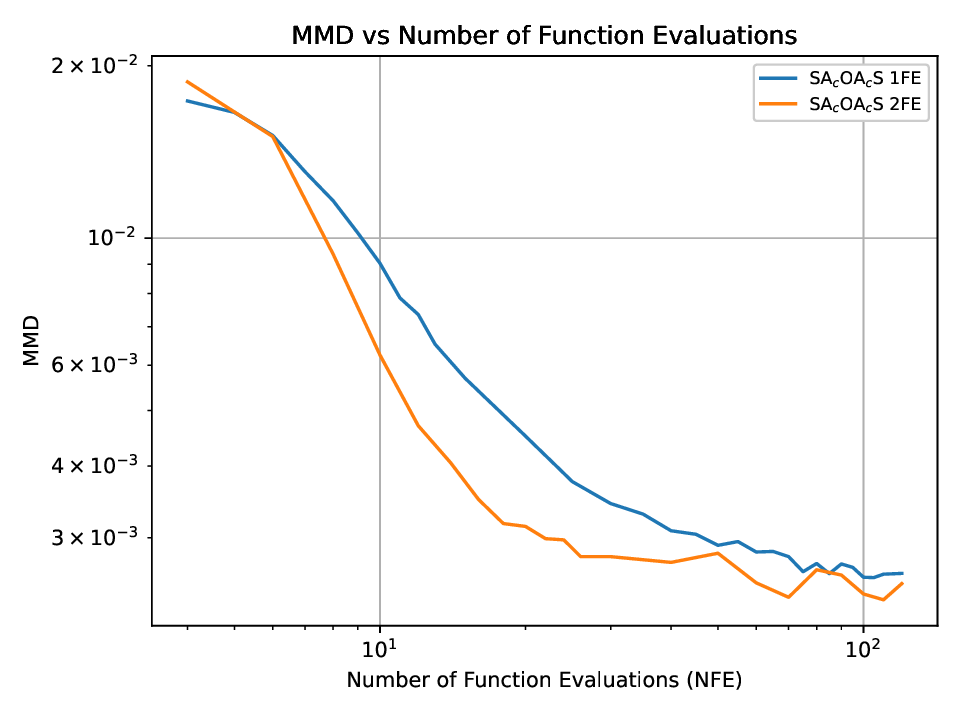}
    \caption{SA$_c$OA$_c$S model comparison for wheel dataset : MMD vs NFE. Recall that $1$ NFE SA$_c$OA$_c$S  is BA$_c$OA$_c$S }
    \label{fig:nfe_wheel_SAOAS_1FE_VS_2FE_2-120}
\end{figure}

\begin{figure}[H]
    \centering
    \includegraphics[width=0.7\linewidth, height = 0.25\textheight]{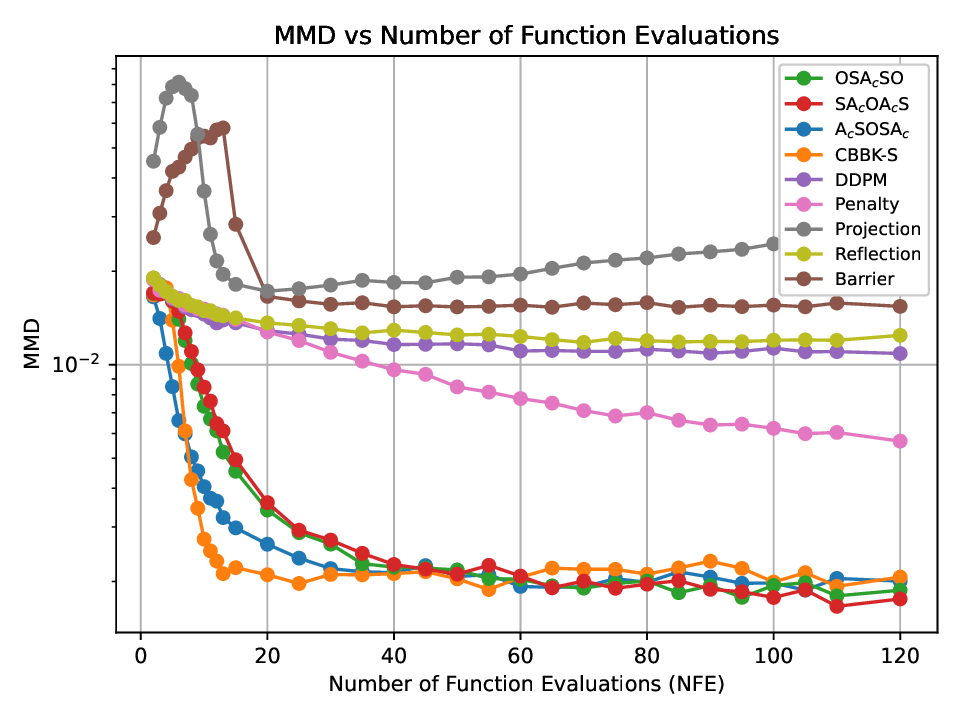}
    \caption{Model comparison of NFE on the wheel dataset. 1FE per iteration}
    \label{fig:nfe_wheel_SAOAS_2-120}
\end{figure}

The models are trained for $4\times10^{4}$ iterations where all the models use 1 FE per reverse step. The average MMD is calculated over $10$ runs. 

\begin{figure}[H]
    \centering
    \begin{subfigure}{0.24\linewidth}
        \centering
        \includegraphics[width=0.9\linewidth]{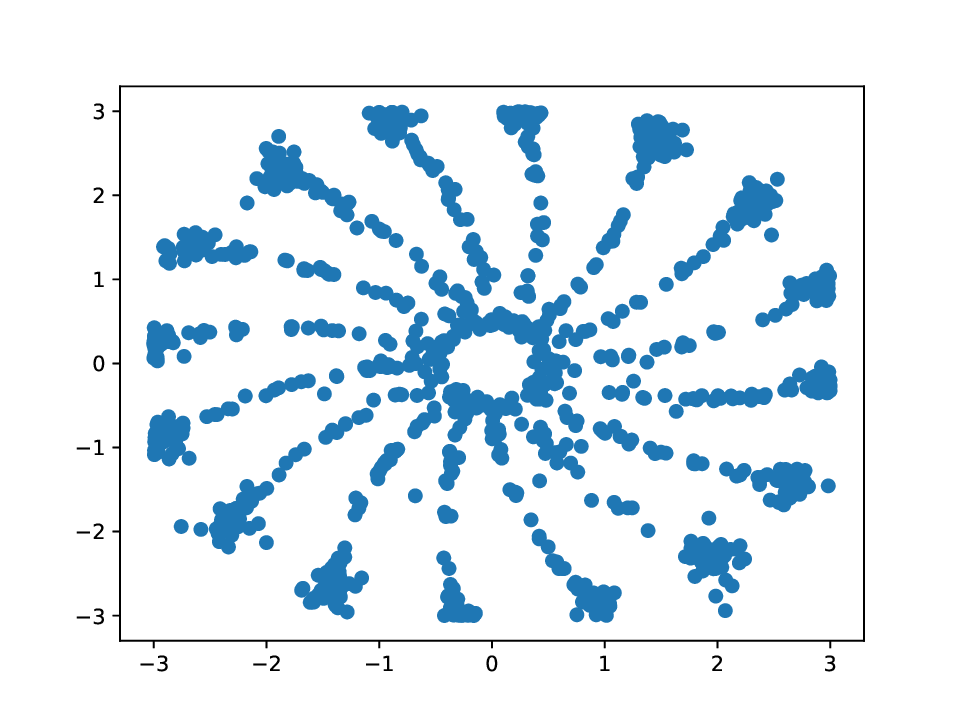}
        \caption{CLD A$_c$SOSA$_c$}
        \label{fig:cld_aboba}
    \end{subfigure}%
    \hfill 
    \begin{subfigure}{0.24\linewidth}
        \centering
        \includegraphics[width=0.9\linewidth]{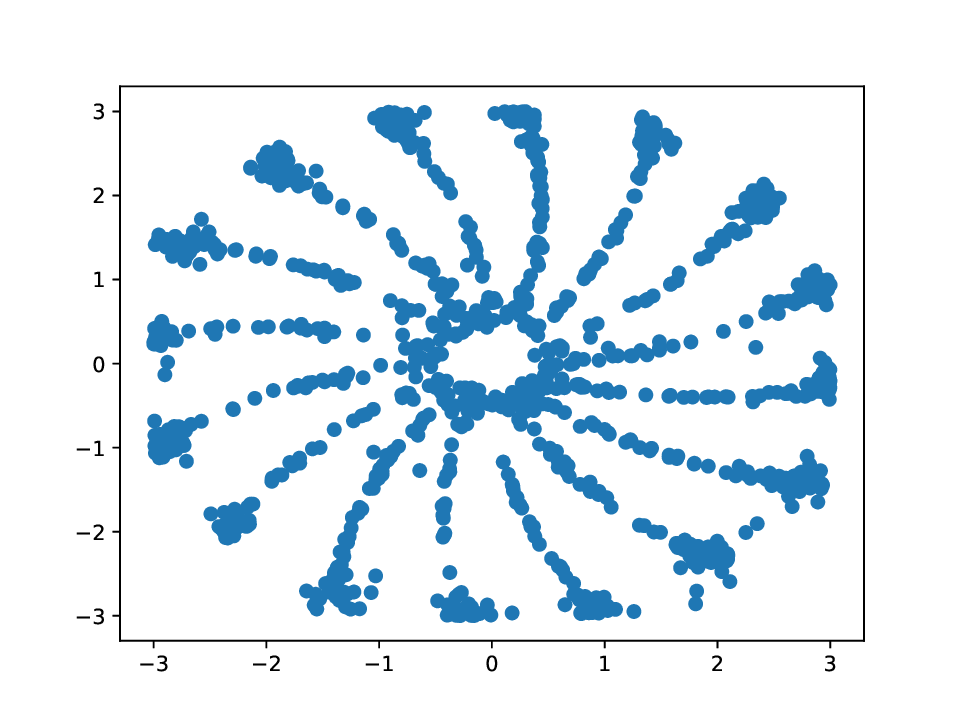}
        \caption{CLD SA$_c$OA$_c$S}
        \label{fig:cld_baoab}
    \end{subfigure}%
    \hfill 
    \begin{subfigure}{0.24\linewidth}
        \centering
        \includegraphics[width=0.9\linewidth]{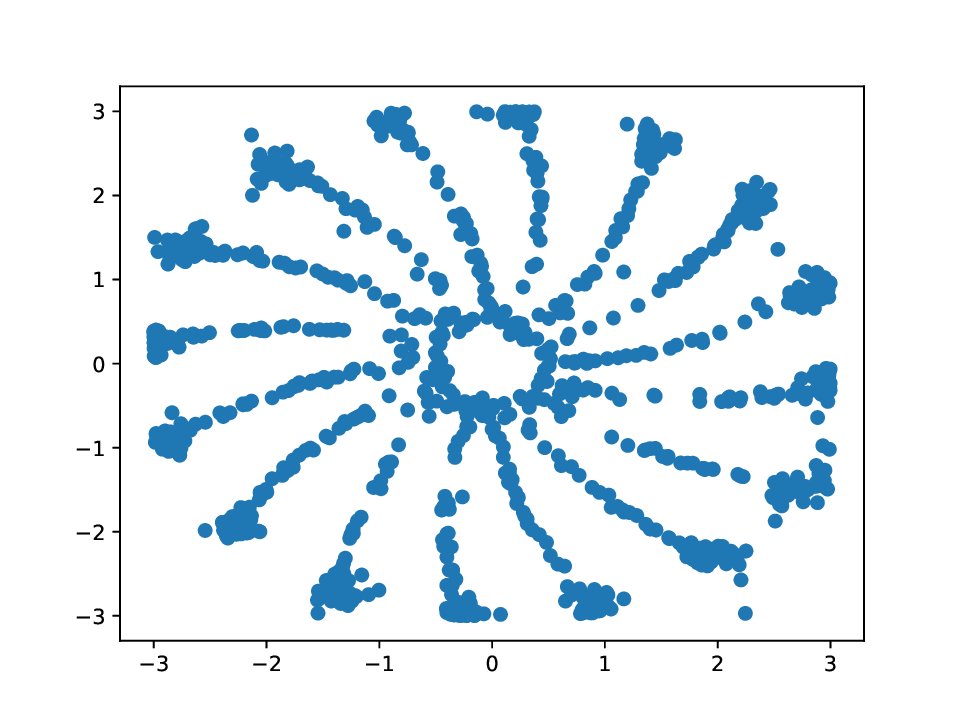}
        \caption{CLD OSA$_c$SO}
        \label{fig:cld_obabo}
    \end{subfigure}%
    \hfill 
    \begin{subfigure}{0.24\linewidth}
        \centering
        \includegraphics[width=0.9\linewidth]{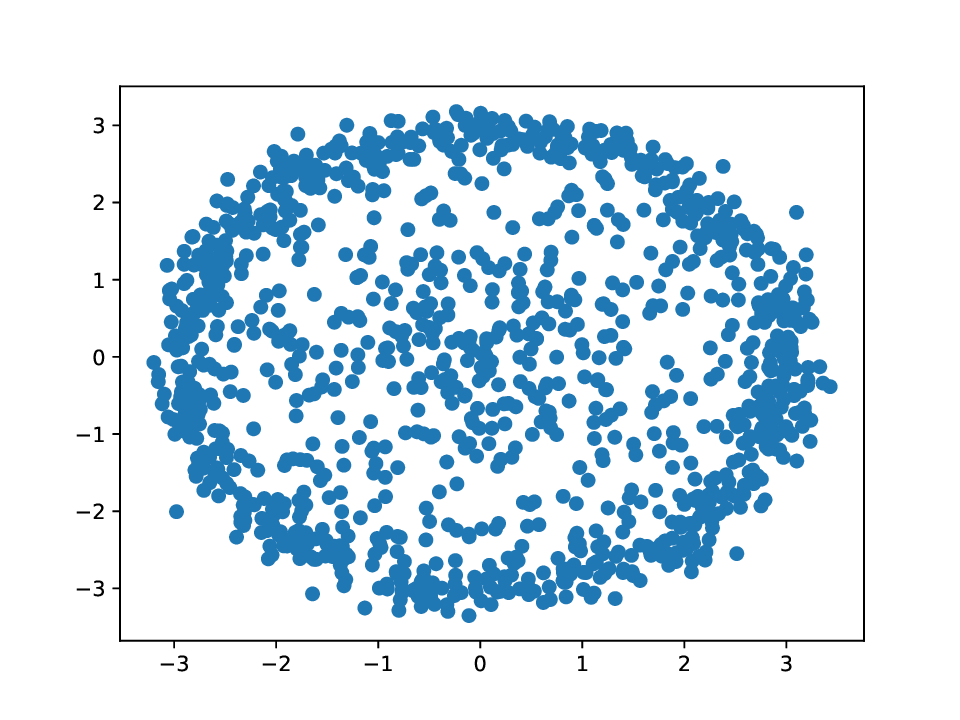}
        \caption{DDPM}
        \label{fig:ou}
    \end{subfigure}
    \caption{Final generated samples; wheel dataset $\gamma=2,b=0$}
    \label{fig:combined}
\end{figure}

\begin{figure}[htbp]
    \centering
    \begin{subfigure}{0.48\linewidth}
        \centering
        \includegraphics[width=0.9\linewidth]{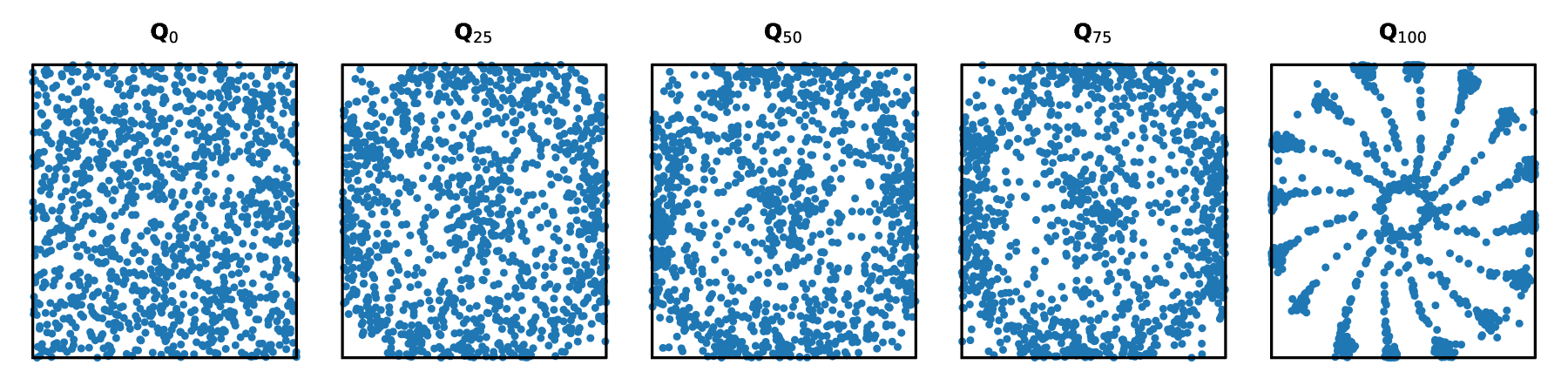}
        \caption{CLD A$_c$SOSA$_c$}
        \label{fig:cld}
    \end{subfigure}%
    \hfill 
    \begin{subfigure}{0.48\linewidth}
        \centering
        \includegraphics[width=0.9\linewidth]{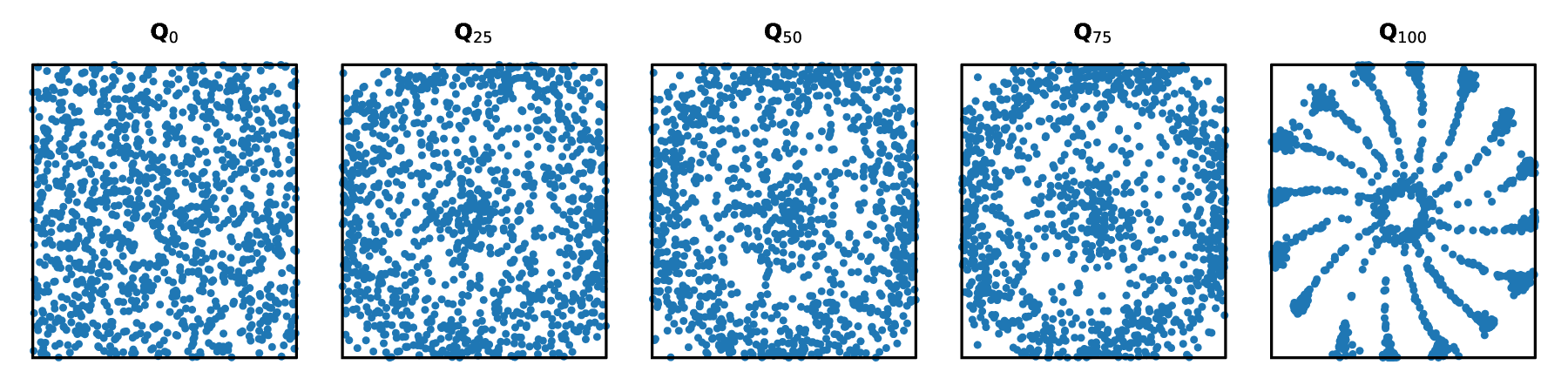}
        \caption{CLD SA$_c$OA$_c$S}
        \label{fig:ou}
    \end{subfigure}

    \vspace{1em} 

    \begin{subfigure}{0.48\linewidth}
        \centering
        \includegraphics[width=0.9\linewidth]{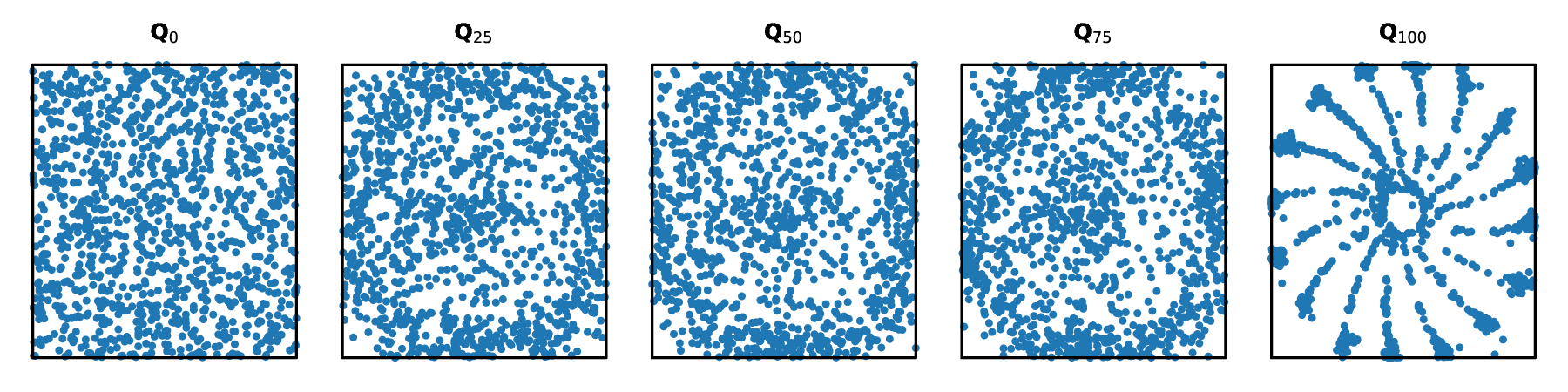}
        \caption{CLD OSA$_c$SO}
        \label{fig:ou}
    \end{subfigure}%
    \hfill 
    \begin{subfigure}{0.48\linewidth}
        \centering
        \includegraphics[width=0.9\linewidth]{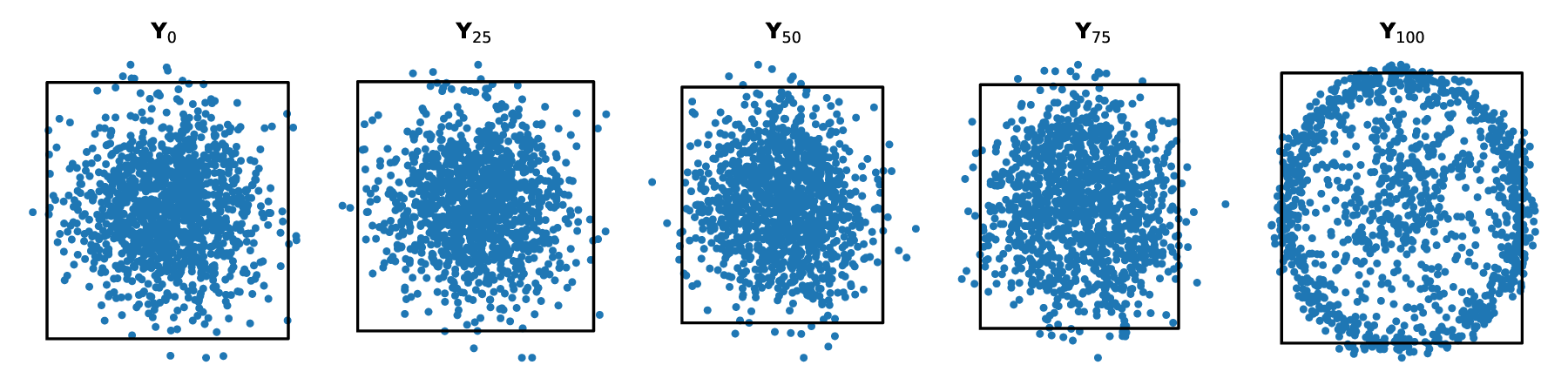}
        \caption{DDPM}
        \label{fig:ou}
    \end{subfigure}
    \caption{Experiment wheel dataset $\gamma=2,b=0$}
\end{figure}

\begin{table}
\centering
\caption{Maximum Mean Discrepancy (MMD) based on 10 sampling runs for the wheel dataset.}
\label{tab:evaluation}
\begin{tabular}{lcc}
\toprule
Model & MMD ($\times 10^{-2}$) & Constraint Violation ($\%$) \\
\midrule
OSA$_c$SO & $\mathbf{0.71} \pm 0.03$ & $0.00 \pm 0.00$ \\
SA$_c$OA$_c$S & $0.72 \pm 0.05$ & $0.00 \pm 0.00$ \\
DDPM & $1.09 \pm 0.03$ & $9.81 \pm 0.76$ \\
A$_c$SOSA$_c$ & $0.73 \pm 0.03$ & $0.00 \pm 0.00$ \\
\bottomrule
\end{tabular}
\end{table}

\begin{table}[h!]
\centering
\caption{MMD and Constraint Violation for $\gamma = 1, b = -x$}
\label{tab:evaluation_gamma1_b-x}
\begin{tabular}{lcc}
\toprule
Model & MMD ($\times 10^{-2}$) & Constraint Violation ($\%$) \\
\midrule
OSA$_c$SO & $0.71 \pm 0.03$ & $0.00 \pm 0.00$ \\
SA$_c$OA$_c$S & $0.72 \pm 0.05$ & $0.00 \pm 0.00$ \\
DDPM & $1.68 \pm 0.05$ & $8.45 \pm 0.40$ \\
A$_c$SOSA$_c$ & $0.73 \pm 0.03$ & $0.00 \pm 0.00$ \\
\bottomrule
\end{tabular}
\end{table}

\begin{table}[h!]
\centering
\caption{MMD and Constraint Violation for $\gamma = 1, b = 0$}
\label{tab:evaluation_gamma1_b0}
\begin{tabular}{lcc}
\toprule
Model & MMD ($\times 10^{-2}$) & Constraint Violation ($\%$) \\
\midrule
OSA$_c$SO & $0.20 \pm 0.02$ & $0.00 \pm 0.00$ \\
SA$_c$OA$_c$S & $0.19 \pm 0.01$ & $0.00 \pm 0.00$ \\
DDPM & $1.68 \pm 0.05$ & $8.45 \pm 0.40$ \\
A$_c$SOSA$_c$ & $0.19 \pm 0.02$ & $0.00 \pm 0.00$ \\
\bottomrule
\end{tabular}
\end{table}

\begin{table}[h!]
\centering
\caption{MMD and Constraint Violation for $\gamma = 2, b = -x$}
\label{tab:evaluation_gamma2_b-x}
\begin{tabular}{lcc}
\toprule
Model & MMD ($\times 10^{-2}$) & Constraint Violation ($\%$) \\
\midrule
OSA$_c$SO & $0.25 \pm 0.01$ & $0.00 \pm 0.00$ \\
SA$_c$OA$_c$S & $0.23 \pm 0.01$ & $0.00 \pm 0.00$ \\
DDPM & $1.68 \pm 0.05$ & $8.45 \pm 0.40$ \\
A$_c$SOSA$_c$ & $0.23 \pm 0.02$ & $0.00 \pm 0.00$ \\
\bottomrule
\end{tabular}
\end{table}

\begin{table}[h!]
\centering
\caption{MMD and Constraint Violation for $\gamma = 2, b = 0$}
\label{tab:evaluation_gamma2_b0}
\begin{tabular}{lcc}
\toprule
Model & MMD ($\times 10^{-2}$) & Constraint Violation ($\%$) \\
\midrule
OSA$_c$SO & $0.19 \pm 0.01$ & $0.00 \pm 0.00$ \\
SA$_c$OA$_c$S & $0.20 \pm 0.02$ & $0.00 \pm 0.00$ \\
DDPM & $1.68 \pm 0.05$ & $8.45 \pm 0.40$ \\
A$_c$SOSA$_c$ & $0.19 \pm 0.01$ & $0.00 \pm 0.00$ \\
\bottomrule
\end{tabular}
\end{table}

\vspace{-4pt}\paragraph{Maze dataset.}

\begin{figure}[H]
    \centering
    \begin{minipage}{0.55\linewidth}
        \centering
        \begin{subfigure}{0.48\linewidth}
            \centering
            \includegraphics[width=0.9\linewidth]{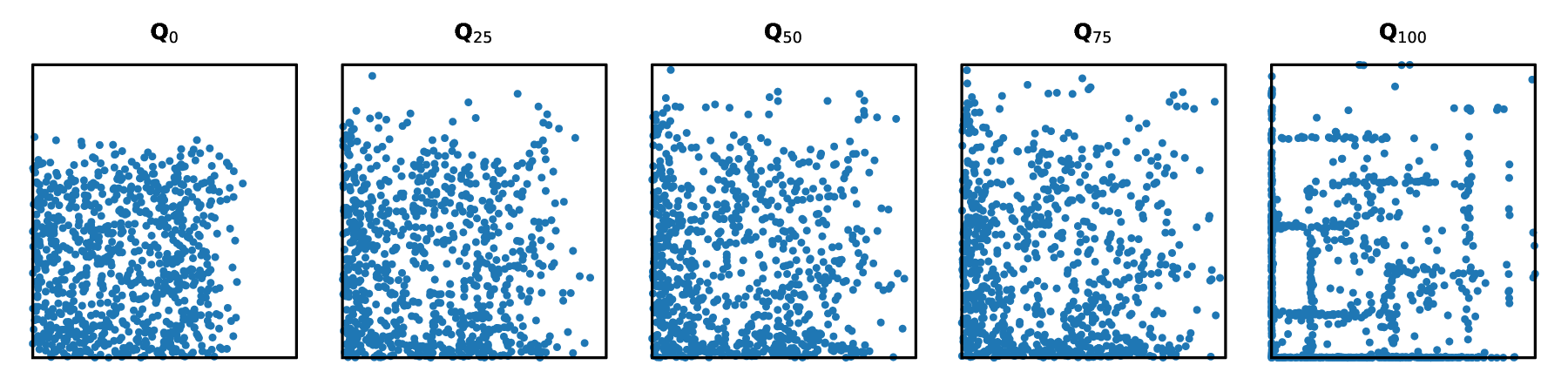}
            \caption{SA$_c$OA$_c$S $\gamma=1,b=-x$}
            \label{fig:cld_left}
        \end{subfigure}%
        \hfill
        \begin{subfigure}{0.48\linewidth}
            \centering
            \includegraphics[width=0.9\linewidth]{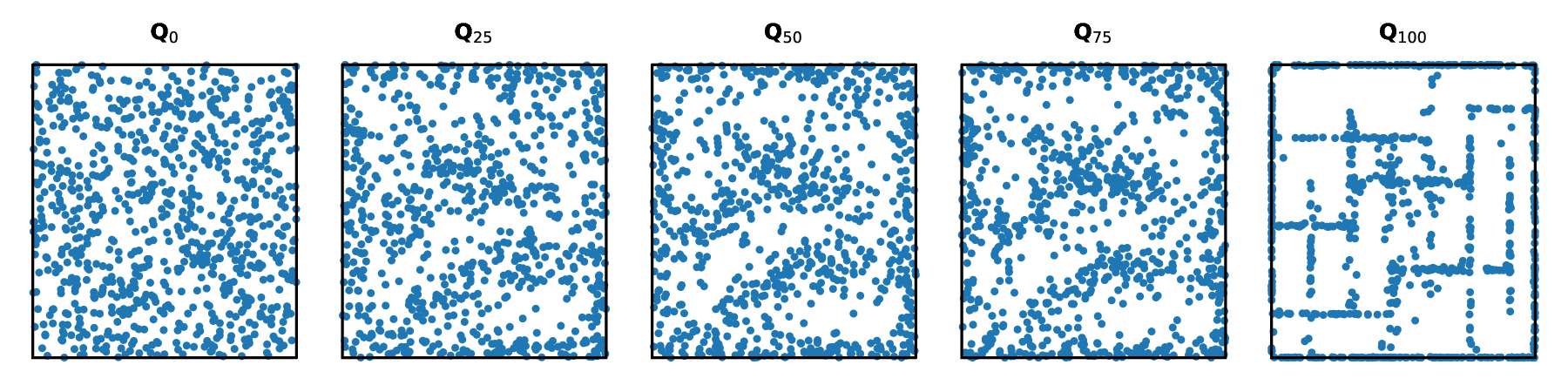}
            \caption{SA$_c$OA$_c$S $\gamma=1,b=0$}
            \label{fig:ou_left_top}
        \end{subfigure}

        \vspace{1em}

        \begin{subfigure}{0.48\linewidth}
            \centering
            \includegraphics[width=0.9\linewidth]{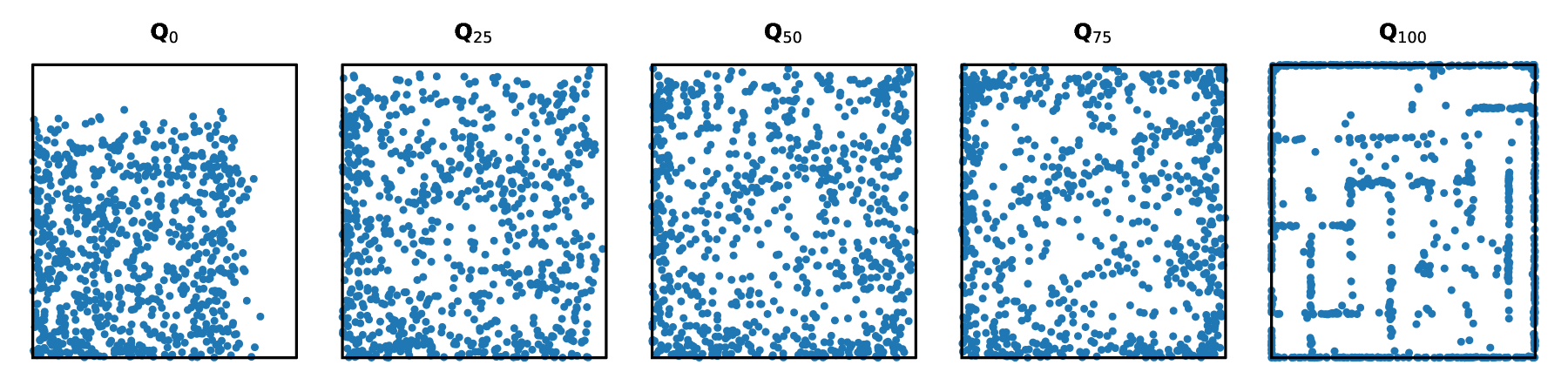}
            \caption{SA$_c$OA$_c$S $\gamma=2,b=-x$}
            \label{fig:ou_left_bottom_left}
        \end{subfigure}%
        \hfill
        \begin{subfigure}{0.48\linewidth}
            \centering
            \includegraphics[width=0.9\linewidth]{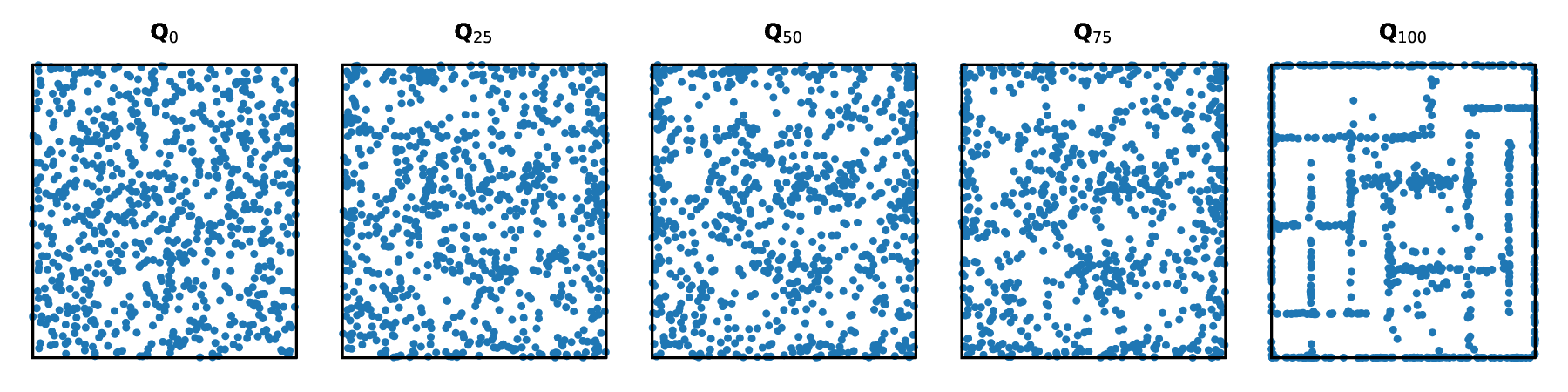}
            \caption{SA$_c$OA$_c$S $\gamma=2,b=0$}
            \label{fig:ou_left_bottom_right}
        \end{subfigure}
        \caption{Generated samples using SA$_c$OA$_c$S (from Figure~\ref{saoas_bbks}) for $k = 20, 40,60,80,100$.}
        \label{fig:maze_Saoas_reverse_step_plots}
    \end{minipage}%
    \hfill
    \begin{minipage}{0.4\linewidth}
        \centering
        \includegraphics[width=0.9\linewidth]{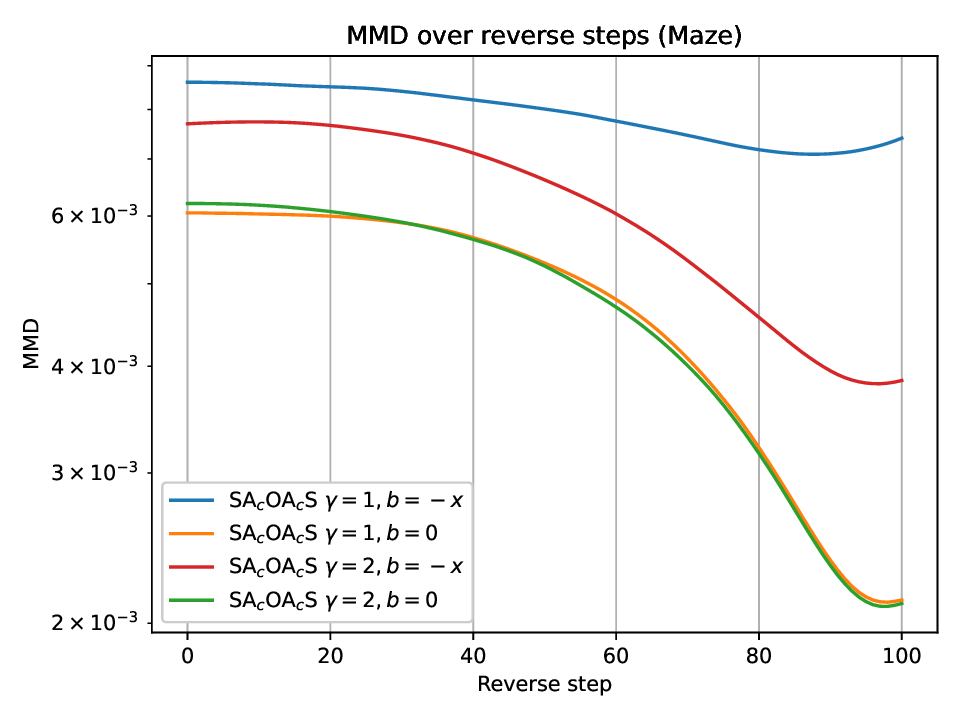}
        \caption{MMD (computed over $10$ runs) comparison for Maze dataset.}
        \label{fig:saoas_maze_mmd_comparison}
    \end{minipage}
\end{figure}

\begin{figure}[H]
    \centering
    \begin{minipage}{0.55\linewidth}
        \centering
        \begin{subfigure}{0.48\linewidth}
            \centering
            \includegraphics[width=0.9\linewidth]{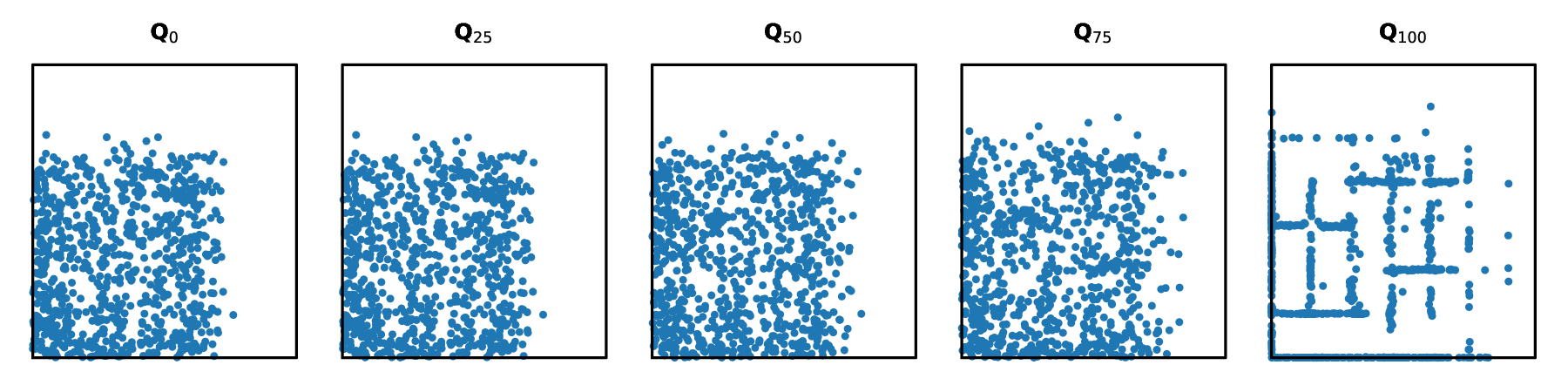}
            \caption{CBBK-S $\gamma=1,b=-x$}
            \label{fig:cld_left}
        \end{subfigure}%
        \hfill
        \begin{subfigure}{0.48\linewidth}
            \centering
            \includegraphics[width=0.9\linewidth]{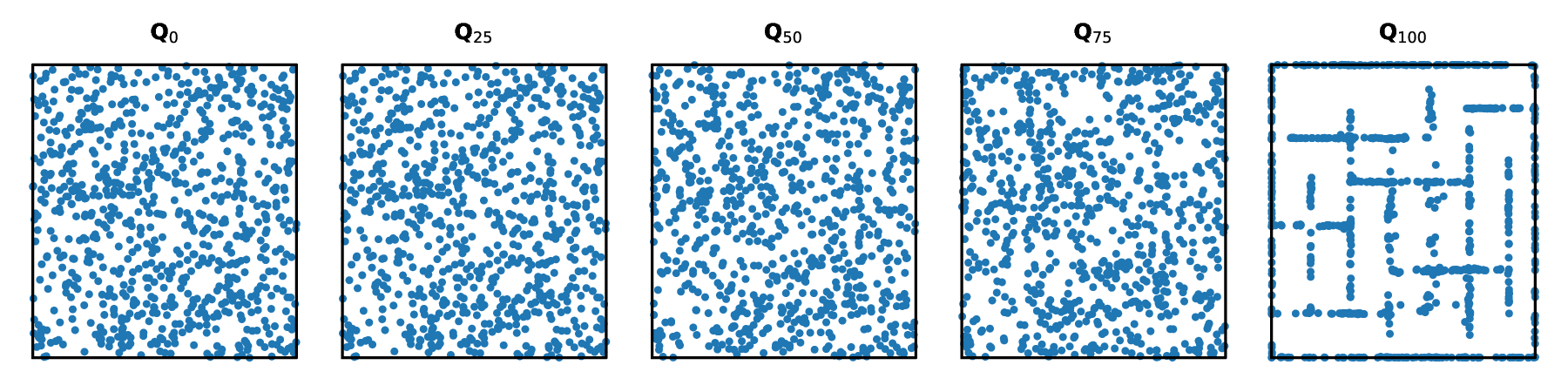}
            \caption{CBBK-S $\gamma=1,b=0$}
            \label{fig:ou_left_top}
        \end{subfigure}

        \vspace{1em}

        \begin{subfigure}{0.48\linewidth}
            \centering
            \includegraphics[width=0.9\linewidth]{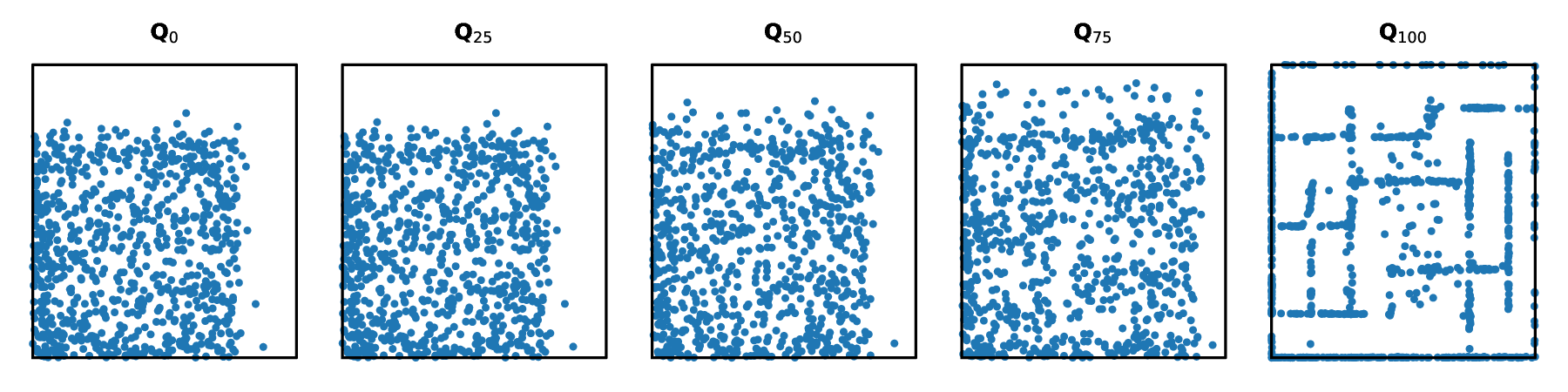}
            \caption{CBBK-S $\gamma=2,b=-x$}
            \label{fig:ou_left_bottom_left}
        \end{subfigure}%
        \hfill
        \begin{subfigure}{0.48\linewidth}
            \centering
            \includegraphics[width=0.9\linewidth]{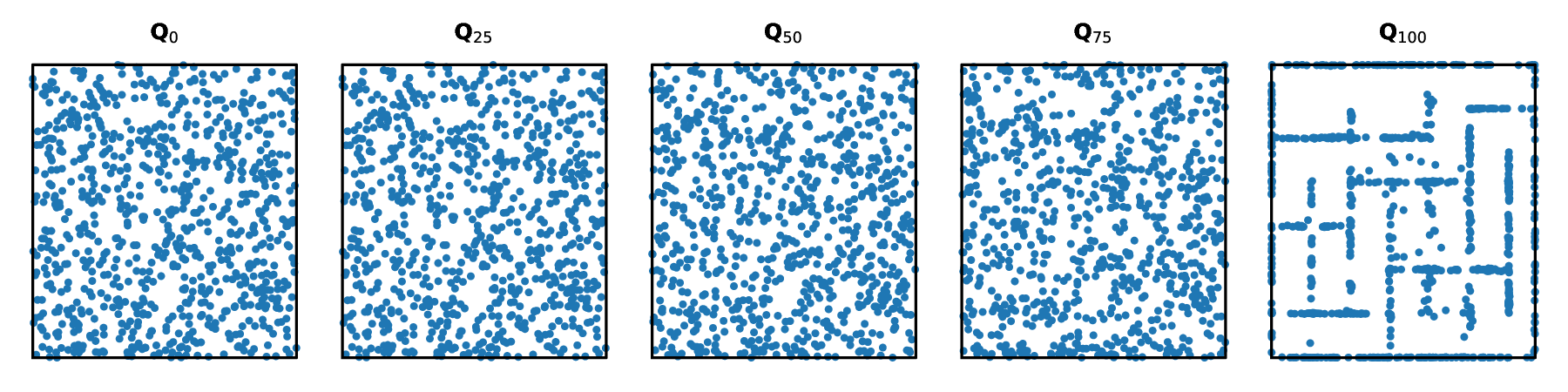}
            \caption{CBBK-S $\gamma=2,b=0$}
            \label{fig:ou_left_bottom_right}
        \end{subfigure}
        \caption{Generated samples using CBBK-S (from Figure~\ref{saoas_bbks}) for $k = 20,40,60,80, 100$.}\label{fig_11_maze_ccbks}
        \label{fig:maze_reverse_step_plots}
    \end{minipage}%
    \hfill
    \begin{minipage}{0.4\linewidth}
        \centering
        \includegraphics[width=0.9\linewidth]{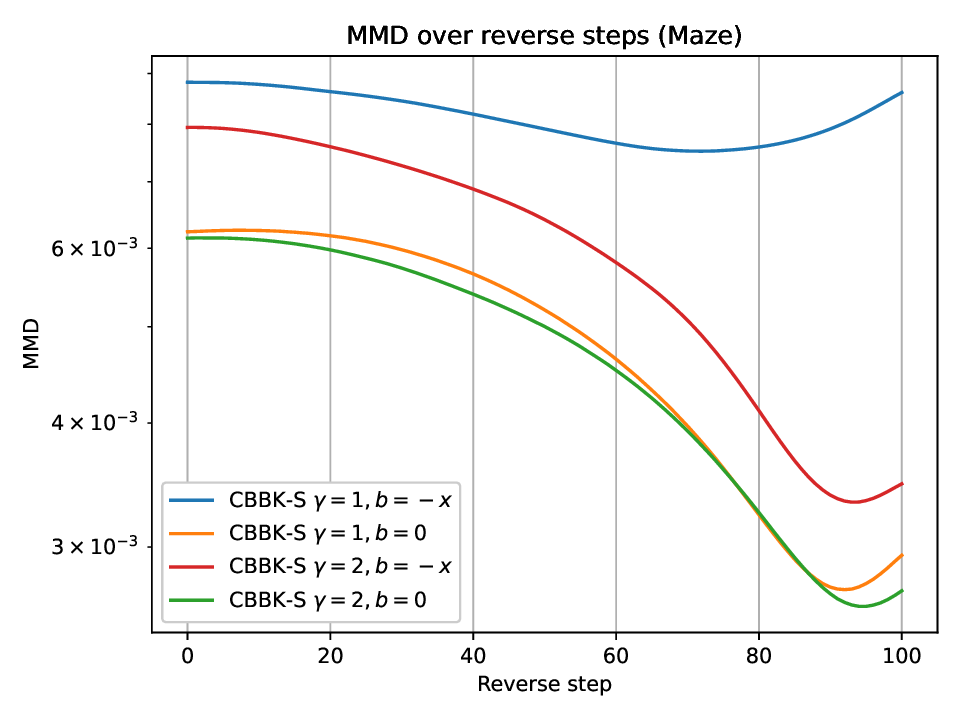}
        \caption{MMD (computed over $10$ runs) comparison for Maze dataset .}\label{fig_12_maze_ccbks_mmd}
        \label{fig:maze_mmd_comparison}
    \end{minipage}
\end{figure}

Here, we compare different choices of $b$ in \eqref{cld_pos_for}-\eqref{cld_vel_for} for a maze looking dataset ($825$ points) in Figures~\ref{fig:maze_Saoas_reverse_step_plots} and ~\ref{fig_11_maze_ccbks}. It is clear from the experiments that the choice of $b=0$ outperforms all other configurations (see Figures~\ref{fig:saoas_maze_mmd_comparison} and~\ref{fig_12_maze_ccbks_mmd}). For implementation, we take $T=1,\;h=0.01$. 

It is clear that the choice of $ b= 0$ gives the best performance.

\paragraph{Flower dataset.}

\begin{figure}[H]
    \begin{subfigure}{0.48\linewidth}
        \centering
        \includegraphics[width=0.9\linewidth]{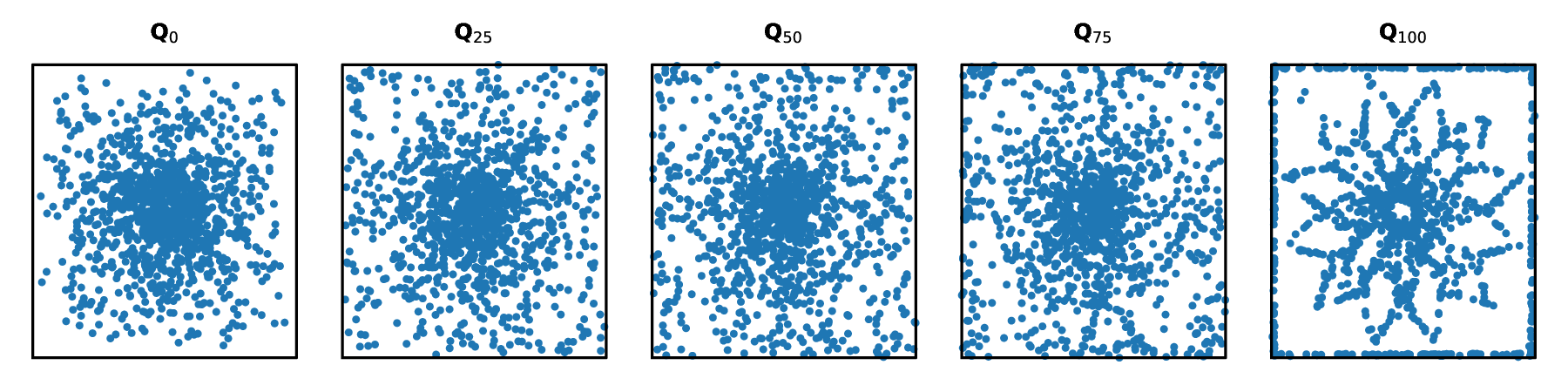}
        \caption{CLD SA$_c$OA$_c$S $\gamma=2,b=-x$}
        \label{fig:ou}
    \end{subfigure}%
    \hfill 
    \begin{subfigure}{0.48\linewidth}
        \centering
        \includegraphics[width=0.9\linewidth]{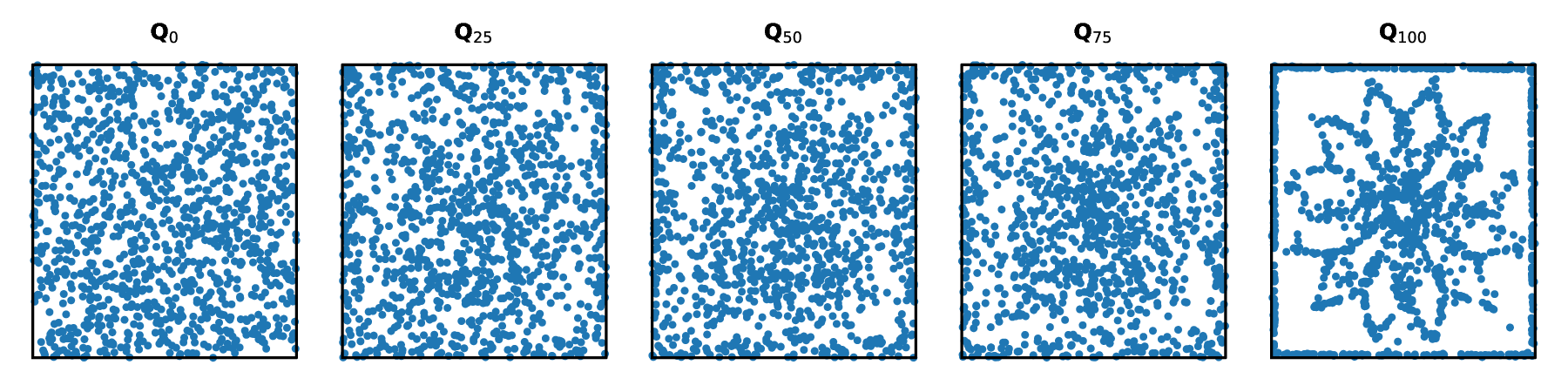}
        \caption{CLD SA$_c$OA$_c$S $\gamma=2,b=0$}
        \label{fig:ou}
    \end{subfigure}
    \caption{Experiment CLD setting using the lotus dataset}
    \label{fig42}
\end{figure}

This dataset lies in the domain $G=\left[-5,5\right]$ which contains $1185$ data points.
The results of the experiments are shown in Table~\ref{tab:evaluation_sacaocas_3_reformatted} as well as in Figures~\ref{fig42}-\ref{fig:combined} for different choices of $\gamma$ and $b$. In this experiment, the choice of $b = -x$ outperforms $b= 0$. 

\begin{figure}[H]
    \centering
    \begin{subfigure}{0.30\linewidth}
        \centering
        \includegraphics[width=0.9\linewidth]{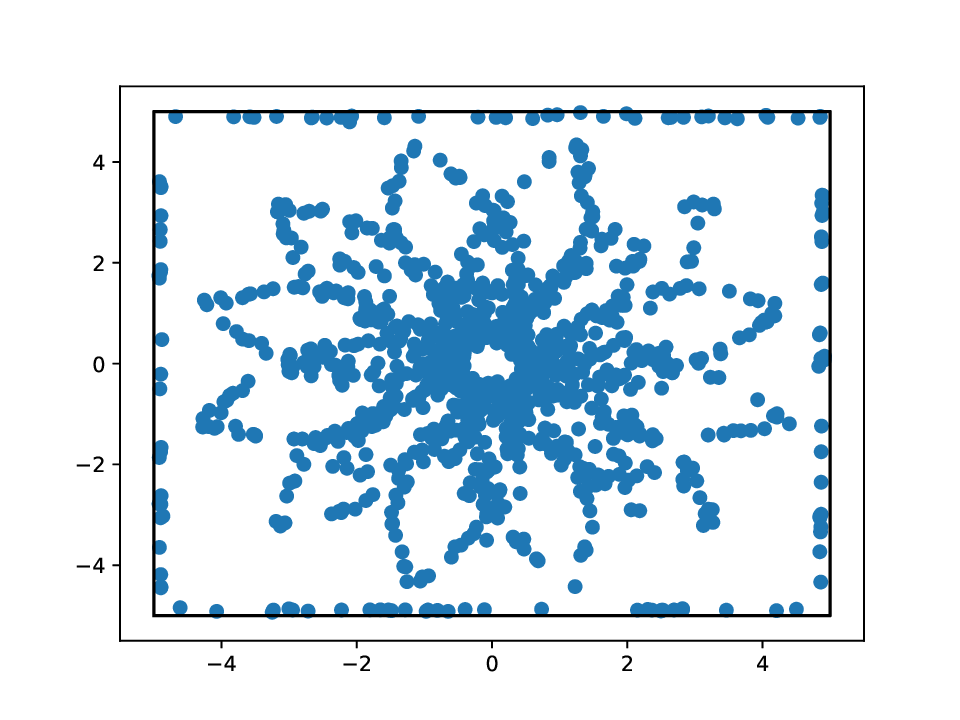}
        \caption{CLD SA$_c$OA$_c$S $\gamma=2,b=-x$}
        \label{fig:cld_obabo}
    \end{subfigure}%
    \begin{subfigure}{0.30\linewidth}
        \centering
        \includegraphics[width=0.9\linewidth]{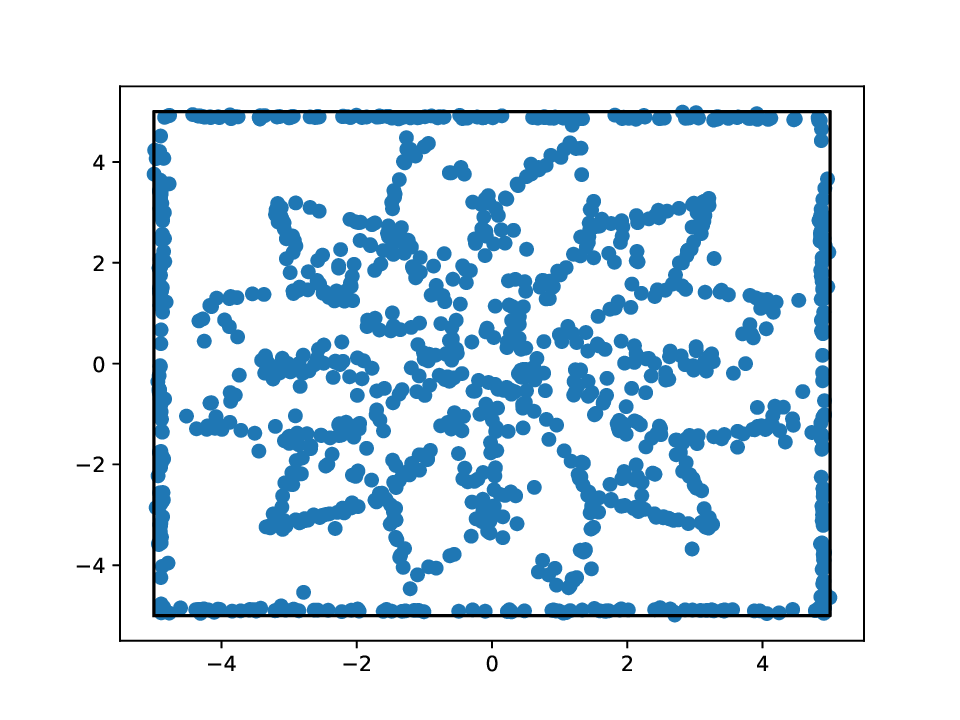}
        \caption{CLD SA$_c$OA$_c$S $\gamma=2,b=0$}
        \label{fig:ou}
    \end{subfigure}
    \caption{Experiment lotus dataset $\gamma=2,b=0$}
    \label{fig:combined}
\end{figure}
\begin{table}[h!]
\centering
\caption{Evaluation Metrics for SA$_c$OA$_c$S under different $\gamma$ and $b$}
\label{tab:evaluation_sacaocas_3_reformatted}
\begin{tabular}{lcc}
\toprule
Model & MMD ($\times 10^{-3}$) & Constraint Violation ($\%$) \\
\midrule
SA$_c$OA$_c$S $\gamma=1,b=-x$ & $1.9 \pm 0.1$ & $0.00 \pm 0.00$ \\
SA$_c$OA$_c$S $\gamma=1,b=0$ & $5.5 \pm 0.1$ & $0.00 \pm 0.00$ \\
SA$_c$OA$_c$S $\gamma=2,b=0$ & $5.4 \pm 0.1$ & $0.00 \pm 0.00$ \\
SA$_c$OA$_c$S $\gamma=2,b=-x$ & $1.1 \pm 0.06$ & $0.00 \pm 0.00$ \\
\bottomrule
\end{tabular}
\end{table}

\subsection{CLD for MNIST dataset} \label{sec_5.3}
\begin{figure}[htbp]
\vspace{-3pt}
    \centering
    \begin{subfigure}{\linewidth}
        \centering
        \includegraphics[width=0.9\linewidth]{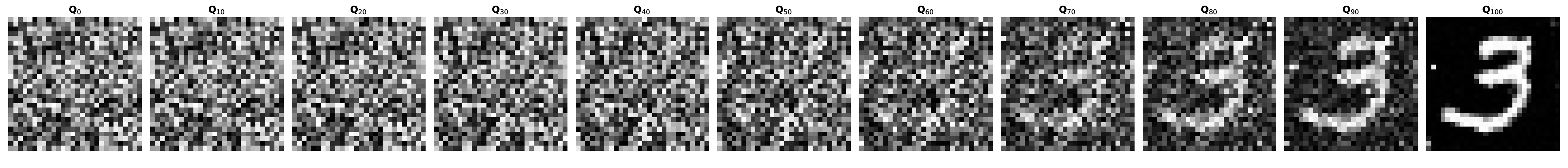}
        \caption{SA$_c$OA$_c$S.}
        \label{fig:cld}
    \end{subfigure}
    \vspace{1em} 
    \begin{subfigure}{\linewidth}
        \centering
        \includegraphics[width=0.9\linewidth]{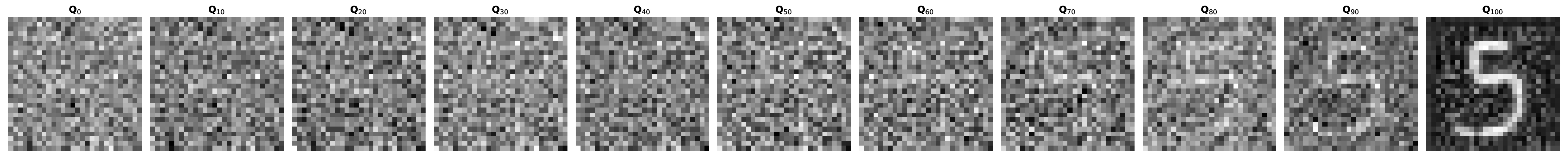}
        \caption{DDPM with clamping.}
        \label{fig:ou}
    \end{subfigure}
    \caption{Sampled digits ($3$ and $5$) generated by the SA$_c$OA$_c$S and DDPM.}
    \label{fig:MNIST_reverse_step_plot}
\end{figure}

The models are trained for $10^5$ iterations with batch size of $64$ on a subset of the MNIST dataset \cite{6296535} containing $18294$ digits from $\{ 1,3,5 \}$ due to computational resources constraints.
Figure~\ref{fig:MNIST_reverse_step_plot} shows two generated samples using DDPM and confined Langevin diffusion model ($G = \left[0,1\right]^{784}$ using SA$_c$OA$_c$S). The number of forward and reverse steps are $100$ with $T=1$ (i.e. $h=0.01$).  Fréchet inception distance (FID) \cite{heusel2018ganstrainedtimescaleupdate} is calculated using the clean-fid library \cite{parmar2021cleanfid} for PyTorch by comparing $15,000$ generated samples by the respective model with $15,000$ samples of the MNIST dataset. The reported FID of the CLD model is $69.85$ and for the DDPM model, it is $185.79$.


\newpage

\newpage 
\appendix
\renewcommand{\thesection}{\Alph{section}}

\doparttoc
\faketableofcontents

\addcontentsline{toc}{section}{Appendices}
\part{Appendices}
\parttoc

\begingroup
\let\clearpage\relax
\endgroup
\section{Technical results}

We make the following assumption on $\partial G$. 

\begin{assumption}
$\partial G$ is $C^{4, \epsilon}$.
\end{assumption}

Also, we will use notation $C$ as a generic constant whose value may change from place to place. 

We use both notation $\rho(t,x, v) $ and $\rho_t(x,v)$  without leading to any confusion. 

\subsection{Proofs related to Score matching loss functions}
\begin{proof}[\textbf{Proof of Proposition~\ref{prop_cld}}]
   \begin{align*}
    \min_{\theta} \text{Loss}(\theta) & =  \min_{\theta}\frac{1}{T}\int_{0}^{T}\mathbb{E}| s_{\theta}(t, X_t, V_t) - \nabla_v \ln \rho(t, X_t, V_t)|^2 \de t
    \\  & 
    =  \min_{\theta}\frac{1}{T}\int_{0}^{T}\big(\mathbb{E}| s_{\theta}(t, X_t, V_t)|^2 -2 \mathbb{E}\la s_{\theta}(t, X_t, V_t) \cdot \nabla_v \ln \rho(t, X_t, V_t)\ra  
    \\ & \quad  +  \E|\nabla_v \ln \rho(t, X_t, V_t)|^2 \big)\de t
    \\   & 
    =  \min_{\theta}\frac{1}{T}\int_{0}^{T}\big(\mathbb{E}| s_{\theta}(t, X_t, V_t)|^2 -2 \mathbb{E}\la s_{\theta}(t, X_t, V_t) \cdot \nabla_v \ln \rho(t, X_t, V_t)\ra  \big)\de t + C,
\end{align*}
where $C>0$ is a positive constant independent of $\theta$.
Using integration by parts, we have
\begin{align*}
  \mathbb{E}\la s_{\theta}(t, X_t, V_t) \cdot \nabla_v \ln \rho(t, X_t, V_t)\ra  &=  \int_{G}\int_{\mathbb{R}^d}  \la s_{\theta}(t, x, v) \cdot \nabla_v \ln \rho(t, x, v)\ra \rho(t,x,v) \de v \de x  
  \\   &  = \int_{G}\int_{\mathbb{R}^d}  \la s_{\theta}(t, x, v) \cdot \nabla_v  \rho(t, x, v)\ra  \de v \de x
  \\  & = -\int_{G}\int_{\mathbb{R}^d}  \diver_{v}\big( s_{\theta}(t, x, v)\big)   \rho(t, x, v)  \de v \de x
  \\  & = -\mathbb{E}_{(X_{t}, V_{t})}\big(\diver_{v}(s_{\theta}(t, X_t, V_t))
\end{align*}
Therefore, we have
\begin{align*}
 \min_{\theta} \text{Loss}(\theta)   &
    =  \min_{\theta} \mathbb{E}_{t\in \mathcal{U}([0,T])}\mathbb{E}_{(X_{t}, V_{t})}\big(| s_{\theta}(t, X_t, V_t)|^2 + 2\diver_{v}(s_{\theta}(t, X_t, V_t)) \big) + C, 
\end{align*}
where the constant $C$ is independent of $\theta$. This completes the proof. 
\end{proof}

\begin{proof}[\textbf{Proof of Proposition~\ref{ref_lossProp_2}}]
 \begin{align}
  \min_{\theta}\text{ Loss}(\theta) &=  \min_{\theta}\frac{1}{T}\int_{0}^{T}\mathbb{E} |s_{\theta}(t, X_t) - \nabla \ln \rho(t, X_t)|^2 \de t  \nonumber 
  \\  & 
  = \min_{\theta}\frac{1}{T}\int_{0}^{T}\mathbb{E} |s_{\theta}(t, X_t)|^2 - 2 \E\la s_{\theta}(t, X_t) \cdot \nabla \ln \rho(t, X_t)\ra +  \mathbb{E} | \nabla \ln \rho(t, X_t)|^2 \de t. \nonumber 
  \end{align}
Due to the divergence theorem and product rule, we have
\begin{align*}
   \int_{G} \nabla \cdot ( s_{\theta} \rho_t(x)) \de x &=  \int_{\partial G} \rho_t(x) \la s_\theta \cdot n(x) \ra \de x, \\ 
  \nabla \cdot (\rho_t(x) s_\theta)    &= \la \nabla \rho_t (x) \cdot s_\theta\ra + \rho_t(x) \diver( s_\theta),
 \end{align*} 
where, for brevity, we have denoted $s_\theta := s_\theta(t,x)$.
This implies
\begin{align*}
    \E\la s_{\theta}(t, X_t) \cdot \nabla \ln \rho(t, X_t)\ra & = \int_{G}\la s_{\theta}(t, x) \cdot \nabla \ln \rho(t, x) \ra \rho(t,x) \de x
    \\  &  = \int_{G}\la s_{\theta}(t, x) \cdot \nabla  \rho(t, x) \ra \de x 
    \\ & = \int_{G} \big(\nabla \cdot ( \rho(t,x) s_{\theta}) - \rho(t, x) \diver (s_{\theta})\big) \de x 
    \\  & = \int_{\partial G}   \rho(t,x) \la s_{\theta} \cdot n(x)\ra \de x - \E \diver (s_{\theta}). 
\end{align*}
Therefore, we obtain
\begin{align}
 \min_{\theta}\text{ Loss}(\theta)   &=  \min_{\theta} \frac{1}{T}\int_{0}^{T} \Big(\mathbb{E}|s_{\theta}(t, X_t)|^2  + 2 \mathbb{E}\diver (s_{\theta}(t, X_t)) \nonumber 
 \\ & \quad  - 2\int_{\partial G}\langle s_\theta(t, x) \cdot n(x)\rangle \rho_t(x) \de x\Big) \de t  +C \nonumber \\ 
  & = \min_{\theta} \mathbb{E}_{t \sim \mathcal{U}([0,T])}\bigg(\mathbb{E}|s_{\theta}(t, X_t)|^2  + 2 \mathbb{E}\diver (s_{\theta}(t, X_t)) \nonumber \\  & \quad 
  - \frac{2}{t}\int_{0}^{t}\int_{\partial G}\langle s_\theta(r, x) \cdot n(x)\rangle \rho_r(x) \de r\bigg) + C, \label{nps_cld_eqn_1}
\end{align}
 where $C $ is independent of $\theta$. 
From \cite[Eqn. 45, p. 283]{gobet2001euler}, we have
\begin{align*}
    \frac{1}{t}\int_{0}^{t} \int_{\partial G} \frac{a(x)}{2}\psi(x) \rho_s(x) \de x \de s =   \frac{1}{t}\mathbb{E}\int_{0}^{t} \psi(X_s) \de L_s,
\end{align*}
where $\psi \in C^{1} (\partial G) $ and $a = n^{\top} \sigma \sigma^{\top} n$ with $\sigma$ being the matrix valued diffusion coefficient. In our setting, we have $\sigma = \sqrt{2} I_d$. Therefore, we get
\begin{align*}
     \frac{1}{t}\int_{0}^{t} \int_{\partial G} \psi(x) \rho_s(x) \de x \de s =   \frac{1}{t}\mathbb{E}\int_{0}^{t} \psi(X_s) \de L_s.
\end{align*}
Choosing $\psi(t,x) = \la s_{\theta}(t, x) \cdot n(x)\ra$ yields
\begin{align}
     \min_{\theta}\text{ Loss}(\theta)  &= \min_{\theta} \mathbb{E}_{t \sim \mathcal{U}([0,T])}\bigg(\mathbb{E}|s_{\theta}(t, X_t)|^2  + 2 \mathbb{E}\diver (s_{\theta}(t, X_t)) \nonumber \\  & \quad 
  - \frac{2}{t}\mathbb{E}\int_{0}^{t}\langle s_\theta(r, X_r) \cdot n(X_r)\rangle \de L_r\bigg) + C.
\end{align}
This completes the proof of Proposition~\ref{ref_lossProp_2}.

\end{proof}
\begin{proof}[\textbf{Proof of Proposition~\ref{ref_lossProp_3}}]
We need the following notation:
\begin{align*}
  n^{\pi}_{k+1} := n(X^{\pi}_{k+1}) =   n(\Pi(X_{k+1}^{'})). 
\end{align*}
Consider the following parabolic PDE:
\begin{align}
    \frac{\partial}{\partial t} u(t,x) + \la b( x) \cdot \nabla_x u(t,x)\ra + \Delta_x u(t,x) = 0, \; (t,x) \in [0,T] \times \bar{G}, 
\end{align}
with non-homogeneous Neumann boundary condition: 
\begin{align}
    \la \nabla_x u(t,x) \cdot n(x) \ra = \psi(t,x), \quad (t,x) \in [0,T] \times \partial G,
\end{align}
and homogeneous terminal condition:
\begin{align}
    u(T,x) = 0, \quad x \in \bar{G}. 
\end{align}
We assume that  the solution $u(t,x)$ exists and is sufficiently smooth (see \cite{sym_euler_bossy_gobet_talay} for similar assumption). The probabilistic representation of solution of the above parabolic PDE is given by
\begin{align}
    u(t_0, x) = \mathbb{E}\int_{0}^{T}\psi(s, X_s) \de L_s,
\end{align}
with $X_0 = x$. 

If we choose $\psi(t,x) = \frac{1}{T} \la s_{\theta} (t,x) \cdot n(x)\ra $, then we have 
\begin{align}
    u(t_0, x) = \frac{1}{T} \mathbb{E}\int_{0}^{T}\la s_{\theta} (s, X_s) \cdot n(X_s)\ra \de L_s,
\end{align}
with $X_0 = x$.

In our scheme (see \eqref{aux_step}), if we replace $\xi_{k+1}^{i}$ with Rademacher distribution i.e. $\xi_{k+1}^{i} \sim \pm 1$ each with probability $1/2$, we have the following results from \cite{leimkuhler2023simplest} (also  see \cite{sharma2021random}):
\begin{align}
    \mathbb{E}(u(t_{k+1}, X_{k+1}^{'}) - u(t_{k}, X_{k}) | X_k) &\leq C h^2, \\
    | u(t_{k+1}, X_{k+1}) + Z_{k+1} - u(t_{k+1}, X_{k+1}^{'}) - Z_k| &\leq C d_{k+1}^{3},  \\
    \E \sum_{k=0}^{N-1}d_{k+1} I_{\bar{G}^c}(X_{k+1}^{'}) & \leq C,
\end{align}
where $C >0$ is independent of $h$ and 
\begin{align}
    Z_{k+1} = Z_{k} + 2 d_{k+1} \psi(t_k, X_{k+1}^{\pi} )I_{\bar{G}^c}(X_{k+1}^{'}),
\end{align}
with $Z_0 = 0$.
In this case, the proof of Proposition~\ref{ref_lossProp_3} follows from Theorem~3.1 in  \cite{leimkuhler2023simplest}.

If we use Gaussian increments for simulation purposes, i.e. $\xi_{k+1}^{i} \sim \mathcal{N}(0,1)$ for all $i=1,\dots,d$ and $k=1,\dots, N-1$, $N = T/h$ (see \eqref{aux_step}), then we proceed as below.

Denoting 
\begin{align}
    Z_{k+1} = Z_{k} + 2 d_{k+1} \psi(t_k, X_{k+1}^{\pi}) I_{\bar{G}^c}(X_{k+1}^{'}),
\end{align}
with $Z_0 = 0$, we have the following approximation of solution:
\begin{align}
    u(t_N, X_{N}) + Z_{N} = \frac{2}{T} \sum_{k= 0}^{N-1}d_{k+1} \la s_{\theta} (t_k, X^{\pi}_{k+1}) \cdot n(X^{\pi}_{k+1})\ra, 
\end{align}
provided the projection $X^{\pi}_{k+1}$ is unique and $X_{k+1} \in G$ for each $k= 0,\dots, N-1$.

Therefore, 
\begin{align*}
    \bigg| \mathbb{E}\frac{1}{T}\int_{0}^{T}&\langle s_\theta(r, X_r) \cdot n(X_r)\rangle  \de L_r 
     - \frac{2}{T} \mathbb{E}\sum_{k=0}^{N -1 }d_{k+1}\la s_{\theta}(t_k, X^{\pi}_{k+1}) \cdot n_{k+1}^{\pi}\ra I_{G^c}(X_{k+1}')\bigg| \\  &  =  |\E (u(t_N, X_N) - u(t_0, X_0))| \\  &  =  \bigg|\E \sum_{k=0}^{N-1} u(t_{k+1}, X_{k+1}) + Z_{k+1}  -u(t_{k}, X_{k}) -Z_{k} \bigg|
    \\  &  = \bigg|\E \sum_{k=0}^{N-1} \E ( u(t_{k+1}, X_{k+1}) +Z_{k+1} -u(t_{k}, X_{k})- Z_{k} |X_k)\bigg|
    \\  & =  \bigg|\E \sum_{k=0}^{N-1} \Big(\E ( u(t_{k+1}, X_{k+1}^{'})  -u(t_{k}, X_{k}) |X_k)  \\  &  \quad + \E( u(t_{k+1}, X_{k+1}) +Z_{k+1}-u(t_{k+1}, X_{k+1}^{'}) -Z_{k}| X_k) \Big)\bigg|
\end{align*}
For Euler scheme, the following one-step error is known result \cite{leimkuhler2023simplest}:
\begin{align}
    \E ( u(t_{k+1}, X_{k+1}^{'})  -u(t_{k}, X_{k}) |X_k) \leq Ch^2,
\end{align}
where $C >0$ is independent of $h$. 

Let $R_0>0$ be sufficiently large number, the choice of which we will discuss below.
We deal now with the error corresponding to reflection from the boundary:
 \begin{align*}
    \E( u(t_{k+1}, X_{k+1}) &+ Z_{k+1} - u(t_{k+1}, X_{k+1}^{'}) - Z_{k} | X_k)  
    \\  & =   \E\big( (u(t_{k+1}, X_{k+1}) -u(t_{k+1}, X_{k+1}^{'}) + Z_{k+1} - Z_{k}) I(|\xi_{k+1}|^2 < R_0) | X_k\big) \\  & 
    +  \E\big( (u(t_{k+1}, X_{k+1}) -u(t_{k+1}, X_{k+1}^{'}) + Z_{k+1} - Z_{k} ) I(|\xi_{k+1}|^2 \geq R_0) | X_k\big)  
\\  &  \leq  C d_{k+1}^{3} I(|\xi_{k+1}|^2 < R_0) + C\mathbb{P}(|\xi_{k+1}|^2 \geq R_0),
\end{align*}
where we have used the fact that $\sup_{x \in G}| u| $ is uniformly bounded due to boundedness of $G$ and the following expansions:
 \begin{align*}
 u(t_{k+1}, X_{k+1})  &= u(t_{k+1}, X_{k+1}^{\pi}) - d_{k+1}\la n(X^{\pi}_{k+1})\cdot\nabla u(t_{k+1}, X_{k+1}^{\pi})\ra \\  & \quad  + (d_{k+1}^{2}/2)D^{2}u(t_{k+1}, X_{k+1}^{\pi})[n(X^{\pi}_{k+1}),n(X^{\pi}_{k+1})] + \mathcal{O}(d^{3}_{k+1}), \\
 u(t_{k+1}, X_{k+1}^{'})  &= u(t_{k+1}, X_{k+1}^{\pi}) + d_{k+1}\la n(X^{\pi}_{k+1})\cdot\nabla u(t_{k+1}, X_{k+1}^{\pi}) \\  & \quad  + (d_{k+1}^{2}/2)D^{2}u(t_{k+1}, X_{k+1}^{\pi})[n(X^{\pi}_{k+1}),n(X^{\pi}_{k+1})] + \mathcal{O}(d^{3}_{k+1}).
 \end{align*}
This splitting of error using $R_0$ is required to ensure the unique projection of $X_{k+1}^{'}$ on $\partial G$. Note that it is possible with a choice of $R_0$ due to Proposition~1 in \cite{sym_euler_bossy_gobet_talay}. 

Note that $d_{k+1}^2 = \dist^2(X_{k+1}^{'}, \partial G) \leq  |X_{k} - X_{k+1}^{'}|^2 = |b( X_{k})h + \sqrt{2 h} \xi_{k+1}|^2  \leq C h |\xi_{k+1}|^2 
 $. 
Therefore, we have
\begin{align*}
    \E( u(t_{k+1}, X_{k+1}) & - u(t_{k+1}, X_{k+1}^{'}) | X_k) \leq C R_0 h d_{k+1} + C\mathbb{P}(|\xi_{k+1}|^2 \geq R_0).
\end{align*}   
This results in the following computations:
\begin{align*}
    \E (u(t_N, X_N) + Z_{N} &- u(t_0, X_0))  =  \E \sum_{k=0}^{N-1} u(t_{k+1}, X_{k+1})  -u(t_{k}, X_{k}) + Z_{k+1} - Z_{k}
    \\  &  \leq Ch + CR_0 h \E \sum_{k=0}^{N-1} d_{k+1} I_{\bar{G}^c}(X_{k+1}^{'})+ C N \mathbb{P}(|\xi_{k+1}|^2 > R_0) 
    \\  &  \leq Ch + C R_0 h \E \sum_{k=0}^{N-1} d_{k+1}I_{\bar{G}^c}(X_{k+1}^{'}) + C \frac{1}{h} \mathbb{P}(|\xi_{k+1}|^2 > R_0)
     \\  &  \leq Ch + CR_0 h  + C \frac{1}{h} \mathbb{P}(|\xi_{k+1}|^2 > R_0), 
\end{align*}    
where we have used Lemma~3.4 from \cite{leimkuhler2023simplest} in the last inequality.
Using Markov's inequality, we have
\begin{align}
    \mathbb{P}(|\xi_{k+1}|^2 \geq R_0) \leq \frac{\mathbb{E}|\xi_{k+1}|^{2p}}{R_{0}^{p}}.  
\end{align}
Take $R_0 = 1/h^{\epsilon + 2/p}$ with arbitrarily small $\epsilon>0$ and arbitrarily large $p>0$, to get 
\begin{align}
    \mathbb{P}(|\xi_{k+1}|^2 \geq R_0) \leq C h^{\epsilon p + 2}.  
\end{align}
Therefore, we get
\begin{align*}
    \E (u(t_N, X_N) +Z_{N} &- u(t_0, X_0))  =  \E \sum_{k=0}^{N-1} u(t_{k+1}, X_{k+1}) + Z_{k+1}  -u(t_{k}, X_{k}) -Z_{k}
    \\  &  \leq Ch + C R_0 h  + C h^{\epsilon p + 1}. 
\end{align*}
Choosing $p = 2/\epsilon$, we have the following result:
\begin{align}
     \E (u(t_N, X_N) + Z_{N} &- u(t_0, X_0)) \leq  Ch^{1-2 \epsilon},
\end{align}
for arbitrarily small $\epsilon$. 

Denoting 
\begin{align}
    \text{Error} := \bigg| \mathbb{E}\frac{1}{t}\int_{0}^{t}&\langle s_\theta(r, X_r) \cdot n(X_r)\rangle  \de L_r 
     - \frac{2}{t} \mathbb{E}\sum_{k=0}^{\lfloor t/h \rfloor -1 }d_{k+1}\la s_{\theta}(t_k, X^{\pi}_{k+1}) \cdot n_{k+1}^{\pi}\ra I_{G^c}(X_{k+1}')\bigg|,
\end{align}
in conclusion, we have the following results:
\begin{itemize}
    \item If $\xi$ is Rademacher random vector and $\partial G \in C^4$, we have
    \begin{align}
        \text{Error} \leq Ch.
    \end{align}
    \item If $\xi$ is Gaussian random vector and $\partial G \in C^4$, we have
    \begin{align}
        \text{Error} \leq Ch^{1-\epsilon},
    \end{align}
    for arbitrarily small $\epsilon >0$.
    \item If $\xi$ is Gaussian random vector and $\partial G \in C^4$ as well as convex, we have
    \begin{align}
        \text{Error} \leq Ch.
    \end{align}
    This is possible because in this case projection of $X^{'}_{k+1} $ on $\partial G$ is unique and no splitting in terms of $R_0$ is required. 
\end{itemize}
This completes the proof. 
\end{proof}

 \subsection{Uniform in time moment bounds}
Using Ito's formula, we have
\begin{align*}
    |V_t|^{2l} & = |V_0|^{2l} + \int_{0}^{t} \la b(X_s) \cdot V_s\ra  |V_s|^{2 l -2}\de s - \gamma \int_{0}^{t} |V_s|^{2 l} \de s +  \sqrt{2\gamma} W_t 
    \\  &  + \sum_{0 \leq s \leq  t}\big(|V_{s^{-}} - 2 \la n(X_s) \cdot V_{s^{-}} \ra n(X_s) I_{\partial G}(X_s) |^{2 l} - |V_{s^{-}}|^{2 l}\big).
\end{align*}
Note that 
\begin{align}
    |V_{s^{-}} - 2 \la n(X_s) \cdot V_{s^{-}} \ra n(X_s) I_{\partial G}(X_s) |^{2 l} = |V_{s^{-}}|^{2 l},
\end{align}
i.e. specular reflection results in conservation of momentum. This implies 
\begin{align*}
    |V_t|^{2l} & = |V_0|^{2l} + \int_{0}^{t} \la b(X_s) \cdot V_s\ra  |V_s|^{2 l -2}\de s - \gamma \int_{0}^{t} |V_s|^{2 l} \de s + \sqrt{2\gamma} W_t, 
\end{align*}
which on taking expectation on both sides gives
\begin{align}
     \mathbb{E}|V_t|^{2l} & = \mathbb{E}|V_0|^{2l} + b_{\max}\int_{0}^{t}   |V_s|^{2 l -1}\de s - \gamma \int_{0}^{t} |V_s|^{2 l} \de s,
\end{align}
where $b_{\max} = \max_{x \in \bar{G}}b(x)$. 
Using Young's inequality, we have
\begin{align}
     \mathbb{E}|V_t|^{2l} \leq \mathbb{E}|V_0|^{2l} - \frac{\gamma}{2} \int_{0}^{t} |V_s|^{2 l} \de s  + \frac{C}{2 \gamma }t
\end{align}
Applying Gronwall-type arguments, we obtain
\begin{align*}
    \mathbb{E}|V(s)|^{2p} \leq C,
\end{align*}
where $C>0$ is independent of $t$.

\subsection{Time reversal of confined Langevin dynamics}

Here, we (formally) derive the reverse-time dynamics $(Q_t, P_t)$ such that $(Q_t, P_t) \sim (X_{T-t}, V_{T-t}) $. Denoting $\tilde{\rho}_{t} := \rho_{T-t}$, we get
\begin{align*}
    &\frac{\partial \tilde{\rho}_t}{\partial t} =  \frac{\partial \rho_{T-t}}{\partial t} = - \frac{\partial \rho_{T-t}}{\partial (T-t)}  \\  & 
    = - \mathcal{L}^{*}\rho_{T-t} (x , v) \\  &
    = \la v \cdot \nabla_x   \rho_{T-t} (x , v)\ra + \la b(x) \cdot \nabla_v   \rho_{T-t} (x , v)\ra - \gamma \la  \nabla_v\cdot (v  \rho_{T-t} (x , v))\ra
\\   &   \quad 
- \gamma \Delta_{v} \rho_{T-t} (x , v).
\end{align*}
Since $ \Delta_{v} \rho_{T-t} (x , v) = \nabla_v \cdot \big( \rho_{T-t} (x , v)\nabla_v \ln \rho_{T-t} (x , v) \big)$,  we have
\begin{align*}
    \frac{\partial \tilde{\rho}_t}{\partial t} &
    = \la v \cdot \nabla_x   \rho_{T-t} (x , v)\ra + \la b(x) \cdot \nabla_v   \rho_{T-t} (x , v)\ra - \gamma \la  \nabla_v\cdot (v  \rho_{T-t} (x , v))\ra
\\   &   \quad 
- \gamma \Delta_{v} \rho_{T-t} (x , v)
\\  & 
 = \la v \cdot \nabla_x   \rho_{T-t} (x , v)\ra + \la b(x) \cdot \nabla_v   \rho_{T-t} (x , v)\ra - \gamma \la  \nabla_v\cdot (v  \rho_{T-t} (x , v))\ra
\\   &   \quad 
-2 \gamma \nabla_v \cdot \big( \rho_{T-t} (x , v)\nabla_v \ln \rho_{T-t} (x , v) \big)  + \gamma \Delta_{v} \rho_{T-t} (x , v)
.
\end{align*}

Identifying the above PDE as Fokker-Planck equation, we have the following form of time-reversed dynamics: 
\begin{align} 
    Q_t &= Q_0 - \int_{0}^{t} P_s \de s, \label{appen_rev_cld_1} \\
      P_t & = P_0 - \int_{0}^t b( Q_s) \de s +  \int_{0}^t  \gamma P_s \de s + 2\int_{0}^t \gamma\nabla_p \ln \rho(T-s, Q_s, P_s)\de s \nonumber \\
       & \quad  + \sqrt{2\gamma}  B_t  - 2 \sum_{0< s \leq t} \langle n(Q_s) \cdot  P_s\rangle n(Q_s)I_{\partial G}(Q_s).\label{appen_rev_cld_2}
\end{align}

In the similar manner, it is not difficult to derive FK-PDE for time-dependent $b$ and $\gamma$. The main issue with the boundary is resolved by the fact that
\begin{align}
    \rho(t, x, v) = \rho( t, x, v - 2\la n(x) \cdot v\ra n(x)),\quad (x, v) \in \{\partial G ; \la n(x) \cdot v\ra > 0 \} 
\end{align}
and using the following form of time reverse SDE with $(Q, R) = (Q , -P)$:
\begin{align*}
    Q_t &= Q_0 + \int_{0}^{t} R_s \de s, \\
      R_t & = R_0 + \int_{0}^t b(Q_s) \de s +  \gamma \int_{0}^t R_s \de s - 2 \gamma \int_{0}^t\nabla_v \ln \rho(T-s, Q_s, -R_s)ds\nonumber \\
       &  + \sqrt{2\gamma}\tilde{B}_t - 2 \sum_{0\leq s \leq t} \langle n(Q(s)) \cdot  R(s)\rangle n(Q(s)),
\end{align*}
where $\tilde{B}_t = - B_t$ is again a Brownian motion.

\section{Experimental Details}
\subsection{Architectures}

\paragraph{Architecture for toy data-sets }
For the toy datasets, a $4$-layered fully connected MLP was used to train the score-network, with $128$ neurons in each hidden layer. Furthermore, time $t$ is appended directly (i.e. no enconding) at the input.
We use SILU activation function except in the output layer. 

\paragraph{Architecture for MNIST}
A U-NET  based on \cite{song2021scorebased} is used for the training of SA$_c$OA$_c$S and DDPM for the MNIST dataset. It uses 4 contracting/expansive layers with a $3\times3$ layered kernel with the number of channels $\left[ 32, 64, 128, 256\right]$ using stride $1$ for the input layer and 2 thereafter. This is followed by a group normalization layer together with the swish activation function. Additionally, gaussian Fourier projection with a embedded dimension of $256$ is used for embedding the timesteps $t$. When the CLD models are trained, the architecture is extended with twice the size in the dimension of the channels.

\subsection{Optimizer and hyper-paremeters}
\paragraph{Exponential Moving Average}
All the models are trained using the Exponential Moving Average (EMA) with  $0.999$ as choice of decay.
\paragraph{Adam optimizer}
We use the Adam optimizer  with learning rate of $5e^{-4}$ without any learning rate (time-step) schedule.

\subsection{Implementation details}
The toy dataset experiments are designed, developed and trained locally using Python version $3.8$ particularly using PyTorch $2.4.1$ and CUDA $12.8$ running on Ubuntu $20.04$. The MNIST dataset is trained on Python version $3.10$ using PyTorch $2.1.0+cu118$ and CUDA $12.7$ running on Ubuntu $22.04$. The toy dataset models are trained using a NVIDIA GeForce RTX 3060 mobile 6GB VRAM whilst the MNIST model was trained on a NVIDIA A40 48GB VRAM.

\subsection{Evaluation metrics}
\paragraph{Maximum Mean Discrepancy}
Maximum Mean Discrepancy (MMD) \cite{JMLR:v13:gretton12a} is used to evaluate the trained models for the toy datasets.
\paragraph{Fréchet inception distance}
For evaluation of the trained SAOAS and DDPM models for MNIST we choose the Fréchet inception distance (FID) \cite{heusel2018ganstrainedtimescaleupdate} for the evaluation. The FID uses the clean-fid library \cite{parmar2021cleanfid} for PyTorch by comparing $15,000$ generated samples by the respective models with $15,000$ samples of the MNIST dataset.
\bibliography{references}
\bibliographystyle{abbrv}
\end{document}